# Moving Objects Analytics: Survey on Future Location & Trajectory Prediction Methods

## Technical Report

HARRIS GEORGIOU, University of Piraeus, Greece
SOPHIA KARAGIORGOU, University of Piraeus, Greece
YANNIS KONTOULIS, University of Piraeus, Greece
NIKOS PELEKIS, University of Piraeus, Greece
PETROS PETROU, University of Piraeus, Greece
DAVID SCARLATTI, Boeing Research and Technology Europe, Spain
YANNIS THEODORIDIS, University of Piraeus, Greece

The tremendous growth of positioning technologies and GPS enabled devices has produced huge volumes of tracking data during the recent years. This source of information constitutes a rich input for data analytics processes, either offline (e.g. cluster analysis, hot motion discovery) or online (e.g. short-term forecasting of forthcoming positions). This paper focuses on predictive analytics for moving objects (could be pedestrians, cars, vessels, planes, animals, etc.) and surveys the state-of-the-art in the context of future location and trajectory prediction. We provide an extensive review of over 50 works, also proposing a novel taxonomy of predictive algorithms over moving objects. We also list the properties of several real datasets used in the past for validation purposes of those works and, motivated by this, we discuss challenges that arise in the transition from conventional to Big Data applications.

CCS Concepts: Information systems → Spatial-temporal systems; Information systems → Data analytics; Information systems → Data mining; Computing methodologies → Machine learning

Additional Key Words and Phrases: mobility data, moving object trajectories, trajectory prediction, future location prediction.

This work was partially supported by the European Union's Horizon 2020 Research and Innovation Programme under Grant Agreements No 687591 (datACRON), No 699299 (DART), No 777695 (MASTER), and No 780754 (Track & Know). Yannis Kontoulis acknowledges a grant received by the State Scholarships Foundation of Greece.

Authors' addresses: H. Georgiou, S. Karagiorgou, Y. Kontoulis, N. Pelekis, P. Petrou, and Y. Theodoridis, Data Science Lab., University of Piraeus, Piraeus, Greece; e-mail: {hgeorgiou|karagior|ikontoulis|npelekis|ppetrou|ytheod}@unipi.gr; D. Scarlatti, Boeing Research and Technology Europe (BRTE), Madrid, Spain; e-mail: david.scarlatti@boeing.com.

# 1. INTRODUCTION

Nowadays, huge amounts of tracking data in the mobility domain are being generated by Global Positioning System (GPS) enabled devices and collected in data repositories; tracked moving entities could be pedestrians, cars, vessels, planes, animals, robots, etc. These datasets constitute a rich source for inferring mobility patterns and characteristics for a wide spectrum of novel applications and services, from social networking applications [5][46] to aviation traffic monitoring [61][67]. During the recent years, this kind of information has attracted great interest by data scientists, both in industry and in academia, and is being used in order to extract useful knowledge about what, how and for how long the moving entities are conducting individual activities related with specific circumstances. The most challenging task is to make this information actionable, by means of exploiting historical mobility patterns in order to gauge how the moving entities may evolve in short- or long- term, whether the individual forecasted movement is typical or anomalous, whether there exists a high probability for congestion in the near future, etc. As a consequence, predictive analytics over mobility data has become increasingly important and turns out to be a 'hot' field in several application domains [4][74][111].

The problem of predictive analytics over mobility data finds two broad categories of application scenarios. The first scenario involves cases where the moving entities are traced in real-time to produce analytics and compute short-term predictions, which are time-critical and need immediate response. The prediction includes either location- or trajectory-related tasks. Short-term location and trajectory prediction facilitates the efficient planning, management, and control procedures while assessing traffic conditions in e.g. road, sea and air transportation. The latter can be extremely important in domains where safety, credibility and cost are critical and a decision should be taken by considering adversarial to the environment conditions to act immediately.

The second scenario involves cases where long-term predictions are important to identify cases which exceed regular mobility patterns, detect anomalies, and determine a position or a sequence of positions at a given time interval in the future. In this case, although response time is not a critical factor per se, it is still crucial in order to identify correlations between historical mobility patterns and patterns, which are expected to appear. Long-term location and trajectory prediction can enhance current plans to achieve cost efficiency or, when contextual information is provided (e.g. weather conditions), it can ensure public safety in different transportation modes (land, sea, air).

In order to present the various aspects of the prediction problem related with moving objects, we discuss what prediction analytics over mobility data would mean, taking an example from the aviation domain. In particular, Fig. 1 illustrates a typical flight, including 'take off', 'top of climb', 'top of descent', and 'touch down' phases. Let us assume that the plane is at the landing process to the arrival airport; typical predictive analytics processes include forecasting about when this plane is expected to touch the ground, whether it should be forced to follow a holding pattern (e.g. due to congestion at the airport), and so on; these are predictions in short-term horizon. On the other hand, let us assume that the plane has just departed; typical analytics processes include forecasting about when and where (in 3-dimensional space) this plane is expected to reach e.g. its 'top of climb' or 'top of descent' phase, whether it is expected to follow the typical motion pattern used by the same flight in the past or, for various reasons, it may need to deviate, and so on; these are predictions in long-term horizon. We could consider similar examples in other use cases as well (cars, vessels, etc.).

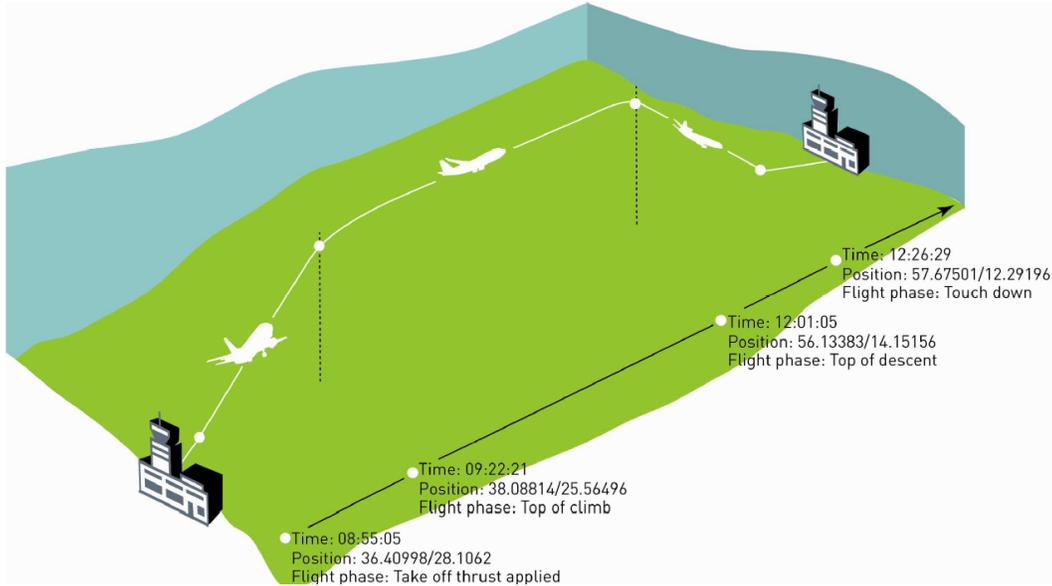

Fig. 1: Predictive analytics over mobility data in practice (example from the aviation domain): assuming that a flight is close to landing at the arrival airport, can we predict when exactly is it expected to touch the ground? Can we predict whether it may be forced to follow a holding pattern before landing? etc. (short-term prediction); assuming that a flight has just departed ('take off' phase), can we predict when (and where) is it expected to reach its "top of climb" and "top of descent" phases? Can we predict whether it may need to deviate, due to various reasons, from the typical flight pattern followed in the past? etc. (long-term prediction). [Figure source: sesarju.eu]

During the last decades, there has been plenty of work on prediction of anticipated movement of objects, including vehicles on road networks, vessels along sea paths, and aircrafts over air corridors. In this work, we provide an extensive literature review of related work so far. Positioning our work in the literature, the most related is a recent survey by Zheng [111], which discusses only a few works about trajectory prediction[1] under the framework of sequential / periodic pattern mining or data privacy. In contrast, our work provides an extensive coverage of the topic including a number of over 50 techniques, spread over the past 20 years. Especially in the aviation domain, there has been an extensive work on prediction models, mainly based on the aircraft characteristics, aerodynamics, and physics laws; a relevant literature survey is found in [67]. In contrast, in our work we focus on data-driven prediction methods, where the historical information available for past flights is used as the main reference for the prediction, in an approach much more big data oriented.

In summary, the main contributions of this paper are as follows:

---

[1] In this section, we use the term "trajectory prediction" informally. In Section 2, we provide formal definitions of the trajectory prediction problem and its variations.

- — we present formal definitions for the different variations of the trajectory prediction problem;
- — we introduce a taxonomy of solutions proposed so far and provide a thorough literature review of state-of-the-art approaches for the problem of interest;
- — we present the properties of the different datasets used in the literature for validation purposes and, motivated by this, we discuss research challenges, especially under the prism of the big data era.

The remainder of the paper is structured as follows. Section 2 provides problem definitions related to prediction over the anticipated movement of a moving object, along with a taxonomy of solutions provided so far. Following this taxonomy, Sections 3 and 4 provide an extensive literature review upon the so-called future location prediction and trajectory prediction methods, respectively. Section 5 summarizes the proposed works and the datasets used for their validation and discusses issues and challenges towards the transition to the big data era; in fact, due to the enrichment of mobility information that is available nowadays as well as the 3 V's (volume, velocity, and variety) challenges. Finally, Section 6 concludes the paper.

## 2. THE MOBILITY PREDICTION PROBLEM

Methodologically, we provide several definitions that cover different aspects of the mobility predictive analytics problem before we survey the state-of-the-art techniques proposed so far.

The *trajectory* of a moving object, it could be a human, an animal, a vehicle, a vessel, a plane, a robot, etc., is defined as a sequence of pairs $<(p_0, t_0), (p_1, t_1), ..., (p_i, t_i), ...>$, where $p_i$ is the location[2] of the object in d-dimensional space (typically, d = 2 or 3, for movement in plane or volume, respectively) and $t_i$ is the time this location was recorded, where $t_i < t_{i+1}$ (i.e., the sequence of pairs $(p_i, t_i)$ is chronologically ordered). In order to simulate the continuous movement of objects, we usually make an assumption of interpolation in-between two consecutive sampled points, $t_i$ and $t_{i+1}$; the most popular is linear interpolation but other formulas may be used as well (B-splines, etc.). A trajectory is called *incomplete* (or *open*) if more locations are expected to arrive, i.e., it is during the evolution of its movement; otherwise, the trajectory is called *complete* (or *closed*). Having these concepts at hand, in the following paragraphs we provide formal definitions of the two most popular prediction-related problems that have been addressed so far, namely *future location prediction* (FLP) and *trajectory prediction* (TP).

In the definitions that we present below, we adopt the following terminology: symbols p and t hold for recorded or given locations and timestamps, respectively, whereas symbols $p^*$ and $t^*$ hold for anticipated (i.e., to be predicted) locations and timestamps, respectively.

---

**Problem definition 1 (Future Location Prediction − FLP)**. *Given the (incomplete) trajectory $<(p_0, t_0), (p_1, t_1), ..., (p_{i-1}, t_{i-1})>$ of a moving object, consisting of its time-stamped locations recorded at past i time instances, and an integer value $j \geq 1$,*

---

[2] In this paper, we use the terms 'location' and 'position' interchangeably; both have the same meaning: a point or range area in the physical 2- (plane) or 3- dimensional (volume) space where an object may be found. In cases where time is considered to be one of the dimensions of the 'space' of interest, it is explicitly mentioned so.

*predict the anticipated trajectory $<(p^*_i, t_i), (p^*_{i+1}, t_{i+1}), ..., (p^*_{i+j-1}, t_{i+j-1})>$ of the object, i.e., object's anticipated locations at the following j time instances.* ∎

**Problem definition 2 (Trajectory Prediction – TP).** *Given the (incomplete) trajectory $<(p_0, t_0), (p_1, t_1), ..., (p_{i-1}, t_{i-1})>$ of a moving object, consisting of its time-stamped locations recorded at past i time instances, and a set C of constraints[3], predict the anticipated trajectory $<(p^*_i, t_i), (p^*_{i+1}, t_{i+1}), ..., (p^*, t^*)>$ of the object, which is consistent with C (note: consistency is not guaranteed to hold for the entire set of constraints).* ∎

---

Practically, in Problem 1, given the 'when' component of the anticipated movement we aim to predict the 'where' counterpart. As an example, and recalling Fig. 1, FLP aims to make an accurate estimation of the next movement of a plane, for instance, the next seconds or minutes of the flight as the plane is approaching at the arrival airport. Usually, FLP is a prediction at short-term horizon.

On the other hand, in Problem 2, given a set of constraints we aim to predict the anticipated movement (both 'where' and 'when' components. As an example, and recalling Fig. 1, TP aims to estimate when and where a plane will be located, from e.g. the take off phase until the landing phase. In contrast to FLP, TP is usually a prediction at long-term horizon

Several variations of TP problem are defined with respect to set $C$. For instance:

- TP problem variation 2a (let us call it *unconstrained TP)*: no constraints at all, i.e., $C = \varnothing$; in this case, we aim to predict anticipated movement without providing any specification.
- TP problem variation 2b (let us call it *destination-constrained TP)*: a single constraint related to the final target, i.e., $C = \{ (p^*$ inside $R) \}$; in this case, we aim to predict anticipated movement with only providing the final destination.
- TP problem variation 2c (let us call it *anchor-constrained TP)*: more than one constraints related to the final as well as intermediate targets, i.e., $C = \{ (p^*_{j'}$ inside $R_j)$, $(p^*_{j'}$ inside $R_{j'})$, $(p^*_{j''}$ inside $R_{j''})$, ..., $(p^*$ inside $R) \}$, we call the intermediate targets $R_j, R_{j'}, R_{j''}, ...$, under the term *anchors*; in this case, we aim to predict anticipated movement with providing a series of intermediate as well as the final destination.

The above problems have been researched in the literature so far, with the majority of the works addressing Problem 1, i.e., the FLP task, where solutions of more general purpose can be proposed, whereas most solutions addressing Problem 2 (with its variations) are application-oriented, e.g. specific for the aviation, the maritime or the urban transportation domain.

In this paper, we survey related work and position it with respect to the above definitions, according to the taxonomy presented in Fig. 2. In particular, Section 3 surveys works on FLP whereas Section 4 focuses on works addressing TP.

---

[3] In the discussion that follows, constraints are upon the spatial dimensions. Nevertheless, constraints could involve time dimension as well. For example, consider an application where the anticipated trajectory should terminate inside a given region $R$ during a given time interval $T$. In order to simplify the discussion, we omit it from the formal definitions above.

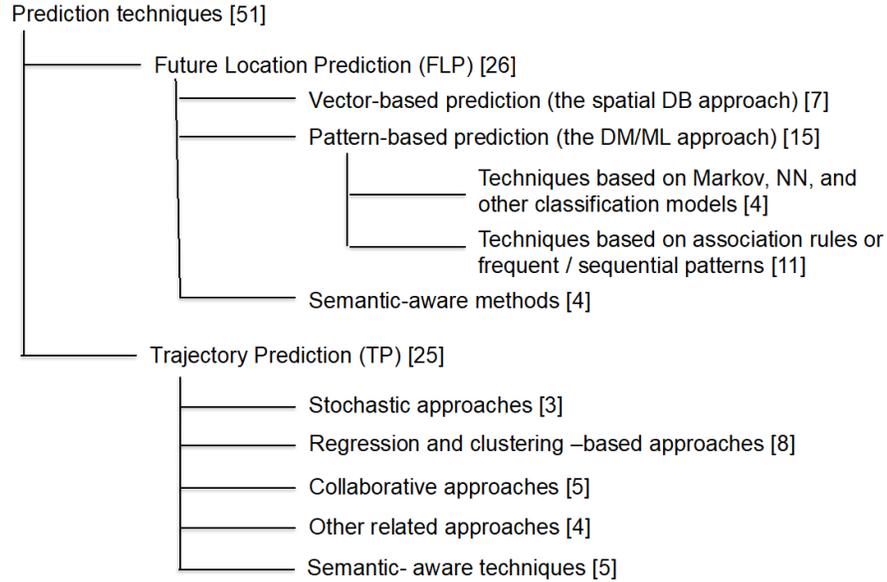

Fig. 2: Taxonomy of FLP and TP methods surveyed in the paper (in brackets, the number of works surveyed per category).

## 3. FUTURE LOCATION PREDICTION

Exhaustive research has been performed so far in order to address the FLP problem (actually, several variations of Problem 1, which was defined in Section 2). The proposals are motivated by (and exploit on) two broad topics: on the one hand, spatial databases (DB) by exploiting on indexing structures organizing movement vectors (hence, vector-based prediction) and, on the other hand, data mining (DM) / machine learning (ML) by exploiting on models and patterns built upon past movements (hence, pattern-based prediction). In the following subsections, we present the most representative proposals from each category.

### 3.1 Vector-based prediction (the spatial DB approach)

This category of techniques includes methods inspired by database management techniques. They take into consideration space and time and predict future locations of moving objects lying in a given time interval using a mathematical or probabilistic model, which aims to simulate the anticipated movement. Actually, most of the methods exploit on the well-known motion function from Physics:

$$p(t) = v \cdot t + p_0 \tag{1}$$

i.e., for each future time instance $t$, the anticipated position $p$ of a moving object is approximately calculated by linear function $p(t)$ above, where $v$ expresses the current velocity and $p_0$ is object's current position. In other words, the above motion function enables to predict locations at any future time $t$ assuming linear extrapolation, as illustrated in Fig. 3. Clearly, this extrapolation may be effective in short-term but cannot be considered a safe prediction in long-term. This short- vs. long- term distinction is a key concept in trajectory prediction and will be discussed in detail along with the surveyed works.

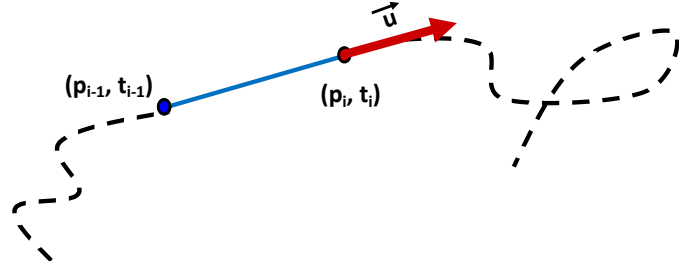

Fig. 3: The future position of a moving object can be considered as the result of a linear motion function taking object's current position and velocity vector into account.

Another important property, in order to follow the rationale behind some of the techniques to be surveyed, is that of *duality*, i.e. the mapping of a hyperplane $h$ from $R^d$ to a point $p$ in $R^d$ and vice-versa. For instance, the linear trajectory:

$$y(t) = v \cdot t + a \qquad (2)$$

of an object moving linearly in 1-dimensional space geometrically constitutes a line in primal plane $(t, y)$, which can be mapped to a point in dual plane $(v, a)$, where $v$ is the velocity and $a$ is the intercept; the so-called Hough-X transform [43] as illustrated in Fig. 4. For objects moving in 2-dimensional space, actually two transformations should be performed, independent for each axis, thus the dual space is 4-dimensional $(v_x, a_x, v_y, a_y)$, and so on. Duality is quite helpful in our discussion, since the (evolving) position of a moving object remains a (stationary) point in dual space as long as it does not modify its velocity (i.e. measure of speed, heading); if this is the case, the mathematical formula representing movement can be efficiently indexed as a point in a spatial access method. Since this family of techniques mainly focuses on the efficient processing (typically, via spatial indexing) of the predicted location of a moving object, they are also called *predictive query processing techniques*.

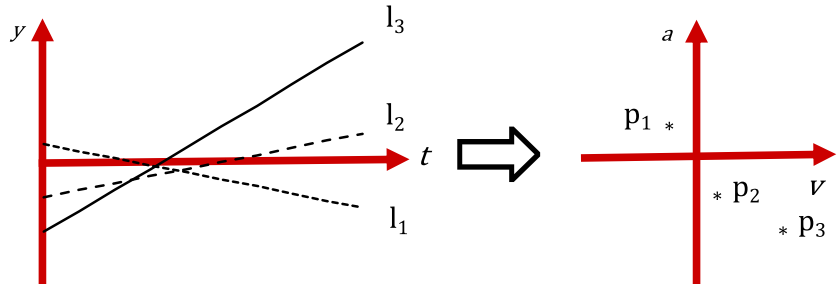

Fig. 4: The concept of duality − mapping the movement of three 1-dimensional objects from line representation in primal $(t, y)$ plane to point representation in dual $(v, a)$ plane. For instance, $l_1$: $y(t) = (-2) \cdot t + 5$ in primal space is mapped to point $p_1$: $(-2, 5)$ in dual space.

In one of the earliest approaches in the field, Tayeb et al. [93] propose an indexing technique based on the PMR (Polygon-Map-Random) Quadtree, a variant of the popular Quadtree data structure for line segments [80]. Inspired by the moving objects spatio-temporal (MOST) data model, according to which the position of a moving object is represented by a function of time [82], the PMR Quadtree variation proposed in [93] stores information about a line segment in every quadrant of the underlying space that it crosses. As dynamic attributes change continuously over time, it becomes difficult to support updates for every change that takes place. This approach addresses the problem of indexing dynamic attributes by adopting a linear time function for each attribute, which describes the way it changes over time and thus enables to predict its value in the future. This is accomplished by introducing Path Computation Algorithm (PCA), which is analogous to Bresenham's algorithm [15] and records in a table $Q[m, n]$ the $m_{th}$ time interval and the $n_{th}$ attribute interval that a moving object crosses. The proposed indexing scheme along with the application of motion function (Eq. 1) above permits the support of two types of queries, namely instantaneous and continuous range queries. In the experimental study presented in [93], it is shown that this methodology achieves high processing performance and reduced disk access cost for both types of queries.

Other techniques are inspired by the popular R-tree family of spatial indexing methods [36][30][56], Modeling of moving object positions as motion functions, on the one hand, address the problem of frequent updates, which could turn out to be curse for the traditional R-tree and its variations, and, on the other hand, enables for predicting tentative future locations of objects. Towards this direction, Kollios et al. [49] propose indexes for mobile objects moving in 1- or 2-dimensional space (also, in the so-called 1.5-dimensional space, where objects move on the plane, though their movement is restricted on using a given collection of routes on the terrain); their indexes exploit on external memory data structures, such as B+-trees [48][19] and kd-trees [11]. In either case, the index maintains dual space-time information, as such it is updated when location changes over time, thus the future location of an object is predicted by taking into consideration a linear function of time and velocity, as it is Eq. 1 above. The proposed indexes exploit linear motion functions for simple movements and they are shown through experimentation to achieve reduced space occupation and better query and update performance as the number of moving objects increases, compared to R*-tree [10].

Applying a similar concept to movement on a road network consisting of road segments, Xu and Wolfson [104] model the future motion plan of an object moving in 2-dimensional space as a trajectory, represented by a polyline in 3-dimensional space. Travel-speed prediction is incorporated whenever an update occurs at a specific road segment, which is then used to update all the trajectories traveling upon this segment. In particular, three alternative methods are proposed: speed update triggered revision (SUTR), query triggered revision (QTR), and query triggered revision with query relaxation (QTR+QR), as an improvement upon the second. The supported future point and range queries assume a 3-dimensional indexing scheme. A point query is implemented in two variations: either retrieves the expected position $p$ of a trajectory at a given (future) time point $t$, or vice-versa retrieves the expected time(s) $t$ when the trajectory will be found at a given position $p$. On the other hand, a range query retrieves the trajectories that are expected to intersect a given region $R$ during a given (future) time interval $[t_1, t_2]$. According to the experimental study presented in [104], it turns out that for a short-term (i.e., up to 15 minutes) travel-speed aware prediction provides more accurate answers than travel-speed unaware prediction and the comparison between QTR+QR and SUTR does not present a clear winner, but the two methods turn out to be suitable for different situations.

Šaltenis et al. [79] propose time-parameterized R-tree (TPR-tree), a spatial index able to support queries over current and future projected positions of moving objects in 1-, 2-, and 3-dimensional space. The moving objects are encoded as points in linear functions of time and use tree bounds to construct bounding rectangles, which are time-parameterized and bound other such rectangles. Technically, TPR-tree is a balanced, multi-way tree with the structure of an R-tree, with the entries in leaf nodes are ($p$, $ptr$) pairs, where $p$ is the position of a moving point and $ptr$ is a pointer to the moving point itself, whereas entries in internal nodes are ($mbr$, $ptr$) pairs, where $ptr$ is a pointer to the root of a sub-tree and $mbr$ is a rectangle that bounds the positions of all moving points or other bounding rectangles in that sub-tree [79]. Actually, the search capability of the TPR-tree index covers three useful types of predictive queries:

– *timeslice query*, where the input of the query is a (fixed) region $R$ at time point $t$ (in geometric terms, a d-dimensional hyper-rectangle);

– *window query*, where the input of the query is a (fixed) region $R$ during time interval [$t_a$, $t_b$] (in geometric terms, a (d+1)-dimensional hyper-rectangle); and

– *moving query*, where the input of the query is an evolving region $R_a \rightarrow R_b$ during time interval [$t_a$, $t_b$] (in geometric terms, a (d+1)-dimensional trapezoid).

The experimental study presented in [79] is based on workloads, which intermix queries and update operations on the index and simulate index usage across a period of time. It is shown there that TPR-tree efficiently supports queries on moving objects, does not degrade severely as time passes, and can be essentially tuned to take advantage of a specific update rate.

Motivated by the TPR-tree, Tao et al. propose an algorithmic framework and analytical cost models for Time-Parameterized predictive queries, which can be applied for dynamic queries and/or dynamic objects [87][88][89], and upon this framework, they later propose the TPR*-tree [91]. Time-Parameterized queries are spatial queries (window, $k$- nearest neighbor and spatial joins) that take into account the changes that may appear in the future due to object movements; as such, the result of a Time-Parameterized query consists of the objects that satisfy the spatial criteria of the query, the expiry time of the result, and the change that causes the expiration of the result. TPR*-tree exploits the characteristics of dynamic moving objects in order to retrieve only those which will meet specific spatial criteria within a given (future) time interval. Each moving object is represented by a Minimum Bounding Rectangle (MBR) along with a Velocity Bounding Rectangle (VBR), which is a vector that expresses the fashion that the object is moving in the 2-dimensional space, according to its velocity at x- and y- axes. In the experimental study presented in [91], TPR*-tree is assessed towards the number of node accesses and the query/update performance compared to the TPR-tree, where it exhibits nearly optimal performance and remains efficient as time evolves. The main idea behind the TPR-tree [79] and its successor, the TPR*-tree [91], is illustrated in Fig. 5.

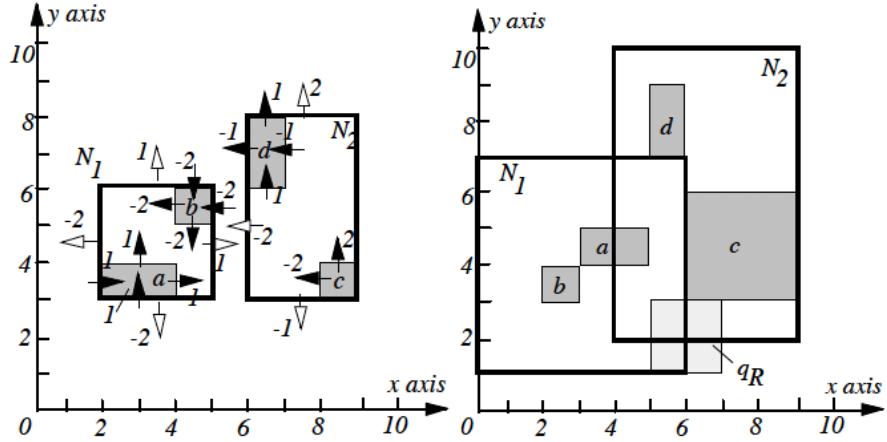

Fig. 5: The main idea behind the TPR-tree: a set of 4 moving points a, b, c, d is grouped in two non-leaf nodes $N_1$ and $N_2$; for each object, either moving point or non-leaf node, the structure maintains its MBR and VBR (arrows and values indicate velocity vectors); the MBR/VBR of a non-leaf entry tightly bounds the MBRs/VBRs of the entries in its child node (timestamp 0, left); after a clock count, each bounding rectangle edge moves according to its velocity; the MBR of a non-leaf entry bounds the MBRs of the entries in its child node but it is not necessarily tight (timestamp 1, right). On the other hand, the TPR*-tree tries to keep them as tight as possible, among other improvements with respect to the TPR-tree. [Figure source [91]]

Motivated by the observation that linear prediction fails not only in long-term (recall Fig. 3) but also in cases where the object's motion is far from being considered as linear (circular movements in a road network ring, etc.), Tao et al. [90] propose Recursive Motion Function (RMF), a prediction mechanism exploiting on a novel indexing scheme, the so-called STP-tree (Spatio-Temporal Prediction tree). In particular, STP-tree is an indexing scheme that incorporates a general framework, which computes different non-linear motion patterns to capture movements of arbitrary modes (linear, polynomial, elliptical, sinusoidal, etc.); the rationale behind the prediction performed is that as diverse motion patterns are met in real world, indexing of unknown motion patterns enables to reflect motion changes over extended periods of time and perform predictive tasks in the distant future. As such, STP-tree it is able to support predictive queries more efficiently than TRP-tree and TRP*-tree, upon which it builds. Technically, RMF exploits the recent past of an object's location and can adapt the prediction according to its individual way of movement. In particular, the method uses as input the actual locations of an object $o$ at the $h$ most recent timestamps and the outcome is the predicted future location of the object at time $t$. Formally [90]:

$$o(t) = C_1 \cdot o(t-1) + C_2 \cdot o(t-2) + \cdots + C_f \cdot o(t-f) \qquad (3)$$

where the location of an object $o$ at time $t$ is a recursive motion function calculated by the object's locations at the $f$ past timestamps using a $d$ x $d$ constant matrix $C_i$ ($f$ is a system parameter, called *retrospect*, and $d$ expresses the movement dimension), or equivalently

$$s_o(t) = K_o \cdot s_o(t-1) \qquad (4)$$

where $s_o(t)$ denotes the motion state of object $o$ at time $t$ and $K_o$ is a constant $d \cdot f \times d \cdot f$ motion matrix for $o$. The motion patterns are classified in known and unknown movements, where in the latter case, the motion matrix varies with the concrete location of the moving object (Fig. 6).

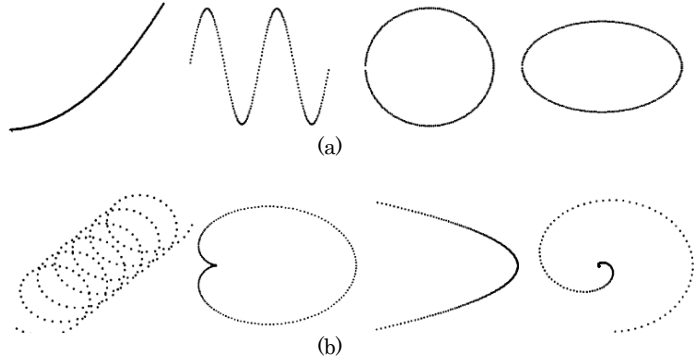

Fig. 6: Examples of motion patterns supported by STP-tree: (a) known (from left to right: polynomial, sinusoid, circle, ellipse) versus (b) unknown patterns (from left to right: spiral, peach, parabola, swirl). [Figures source: [90]]

In the experimental study presented in [90], the framework is evaluated for its expressiveness power using known (polynomial, sinusoidal, circular, elliptical) and unknown movements (spiral, peach, parabola, swirl) and the STP-tree turns out to significantly outperform TPR*-tree in terms of efficiency and effectiveness.

As another alternative to TPR-tree, Jensen et al. [44] propose $B^x$-tree, a solution that exploits on the (ubiquitous) $B^+$-tree [48][19]. The locations of moving objects are modeled and indexed in $B^x$-tree as linear functions coupled with the time they are updated, thus they are represented by timestamped vectors. A two-steps partitioning is performed: first, the time dimension is partitioned into intervals, where the duration of an interval is approximated by the maximum duration in-between two updates of any object location; second, each interval is partitioned into $n$ equal-length sub-intervals, termed *phases*, where $n$ is determined based on minimum time duration within which each object issues an update of its position. Each phase is assigned the time point it ends as a *label timestamp*, and a label timestamp is mapped to a partition. The value indexed by $B^x$-tree is a combination of a moving object's partition number, the result of applying a space-filling method to the object's position (so that a 2-dimensional point is mapped to a 1-dimensional number) [63], and the label timestamp of its phase. In order for $B^x$-tree to support predictive queries (range, continuous, and $k$ nearest neighbor), the search algorithms consider snapshots of moving objects and use query-window enlargement by applying a time parameterized region function. In the experimental study presented in [44], it is demonstrated that $B^x$-tree using Peano ($\varnothing$ or, more preferably, Hilbert space-filling curve is efficient and robust, significantly outperforming TPR-tree regarding I/O cost and query response time.

A more recent approach for addressing such tasks comes from the area of Predictive Queries (PQ) [122][117], which is one of the most exciting research topics in spatio-temporal data management. In many location-based services, including traffic management, ride sharing, targeted advertising, etc, there is a specific need to detect and track mobile entities within specific areas and within specific time frames.

In Range Queries (RQ), the task is focused on identifying POIs and mobility patterns related to the current locations of moving objects. Instead, Predictive Range Queries (PRQ) address the same task but for future time frames. This is a typical use case in aviation, when one or more airplanes need to be checked in some spatial context in the future, e.g. for proximity (collision avoidance), scheduling (takeoff/landing), airspace sectorization (avoid overload and/or delays), etc.

In the context of PRQ and most commonly in the RQ task, various approaches can be used for checking arrivals/departures of airplanes to/from specific regions (e.g. airports), including optimized k-nearest-neighbour (k-nn) variants that employ spatio-temporal index trees. Similarly, a reverse k-nn query can be used to detect moving objects that are expected to have the query region as their nearest neighbour, e.g. for assigning airplanes to their "nearest" service point (ATC hub). Indexing can be implemented by very efficient data management structures, such as R-trees (time-parameterized, a.k.a. TPR/TPR*-trees), variants of B-trees, kd-trees, Quad-trees, etc [122][123]. The predictive model itself can be linear or non-linear and it is most commonly based on historical data in the same spatio-temporal context, in the short- or the long-term w.r.t. time frame. The uncertainty of the prediction is addressed by either model-based approaches, which determine a representative model for the underlying mobility pattern, or pure data-driven approaches, which "learn" and index movements from historic data [118].

Another important aspect in TP for the aviation domain is the ability to employ such models in streaming data, i.e., using "live" sources of mobility data as they become available. This task can also be addressed by PRQ approaches, more specifically the continuous PRQ algorithms. The difference between a "snapshot" predictive query and a continuous one is that the second can be continuously re-evaluated with minimal overhead and optimal efficiency. As an example, the Panda system [120][116], designed to provide efficient support for predictive spatio-temporal queries, offers the necessary infrastructure to support a wide variety of predictive queries that include predictive spatio-temporal range, aggregate (number of objects), and k-nn queries, as well as continuous queries. The main idea of Panda is to monitor those space areas that are highly accessed using predictive queries. For such areas, Panda pre-computes the prediction of objects being in these areas beforehand.

Similar approaches from other contexts, such as the iRoad [119], which is employed for tracking vehicles in urban areas. More specifically, the system supports a variety of common PQs including point query, range query, k-nn query, aggregate query, etc. The iRoad is based on a novel tree structure named reachability tree, employed to determine the reachable nodes for a moving object within a specified future time T. By employing spatial-aware pruning techniques, iRoad is able to scale up to handle real road networks with millions of nodes and it can process heavy workloads on large numbers of moving objects. Since flight routes of civilian and cargo flights are also constrained by submitted flight plans and ATC instructions, similar road-based approaches [121][123] can be adapted for the aviation domain.

In the context of scalability and the Big data aspect, there are very recent and promising approaches such as the UITraMan [124], which addresses the scalability, the efficiency, the  persistence and the extensibility of such frameworks. More specifically, it extends Apache Spark w.r.t. data storage and computing by employing a key-value store and enhances the MapReduce paradigm to allow flexible optimizations based on random data access. Another approach for PQs in Big data is presented by Panda* [125], which is a scalable and generic enhancement of Panda [120], applied to traffic management. More specifically, Panda* is a generic framework for supporting spatial PQs over moving objects, introducing prediction function when there is lack of historic data, isolation of the prediction calculation from the query processing and control over the trade-off between low latency responses and use of computational resources. For both UITraMan and Panda*, experimental results on large-scale real and synthetic data sets in other domains, which include comparisons with the state-of-the-art methods in this area, show promising results and hints of successful application to the aviation domain too.

It should be noted that there are also other types of PQs, more advanced than the ones presented above, such as the predictive pattern queries (PPQ), which check conditions muc more complex than simple presence or not of a moving object within a specific spatio-temporal frame. Such advanced PPQs can be considered as a link between data management and data analytics, which can be very valuable in the context of the aviation domain.

## 3.2 Pattern-based prediction (the DM / ML approach)

This category of prediction approaches exploits on relevant DM or ML methods, considering the prediction task as an instantiation of classification (e.g. Markov models) or an application of frequent / sequential pattern mining, hence the two subsections that follow. An important difference with respect to the methods of the previous section is that, in this case, the methods build upon the history of movements, not only of the object of interest, but also of the other objects moving in the same area; therefore, they are able to build models about the transition from place to place and use them for addressing the FLP task.

### 3.2.1 Techniques based on Markov, NN, and other classification models

Discrete Markov and other classification models have been extensively used to address the future location prediction problem. The main idea is illustrated in Fig. 7: supposed that we build e.g. a Markov chain model of states *a, b, c, d*, etc., based on the history of movement recorded so far, then the next state to be visited (and the probability that this is expected to happen) is a task to be answered by the model itself.

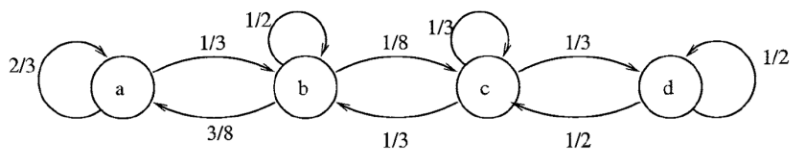

Fig. 7: Markov chain model for movement profile; states correspond to places (of desired spatial granularity) and arrows correspond to transitions (and respective probabilities) from a state to its next state [Figure source: [12]]

In one of the earliest works in this field, Bhattacharya and Das [12] propose LeZi-Update, an adaptive on-line algorithm exploiting on mobility tracking data, characterized as a stochastic process, in a Personal Communication Systems (PCS) network to infer knowledge from subscribers' profiles. This is realized by building and maintaining a dictionary over individual user's path updates, which is trained on user mobility characteristics with optimal message exchange. Their approach relates the complexity of mobility tracking to the location uncertainty of the mobile quantified through the Shannon's *entropy* measure [81], which is used in order to express the least amount of message that needs to be exchanged so that the exact location is known. The dictionary incorporates the technique of lossless compression by deploying the Lempel-Ziv algorithm family [114] in the update scheme and achieves low update and paging costs. The update algorithm captures the sampled message and processes it in chunks. When it finally triggers an actual update, it reports in an encoded form the whole sequence of sampled symbols withheld since the last reporting. As a consequence, the encoded user's movement history becomes distinct. The latter results in incrementally building the dictionary in a global fashion by parsing conditional probabilities estimates of movement histories and providing a robust mechanism able to predict group behavior. The predictive tasks are supported through the dictionary, which records the movement history, enables to learn the mobility profile of users and predict their mobile cell. In the experimental study, it is shown that the proposed approach is efficient for update, paging and planning operations and enables for the provision of quality of service (QoS) and better bandwidth management in wireless data networks.

Ishikawa et al. [42] introduce an algorithm, which extracts mobility statistics from indexed spatiotemporal datasets for interactive analysis of huge collections of moving object trajectories. Their approach structures the trajectories in a R-tree index [36][56] and computes mobility statistics, termed as Markov transition probabilities, by using the index. The statistics are produced by efficiently organizing the target space, i.e., the trajectories of moving objects, and the Markov chain model in a cell-based form. The transition probabilities enable to calculate future cells by using state information from the current cells. A reduction step is applied which assigns the cells expressing the spatial constraints to enumerated groups of objects each of which satisfies a kind of temporal constraints. The latter solves a constraint satisfaction problem (CSP) from the root toward the leaves of the index defined by the spatiotemporal constraints. The CSP-based algorithm uses the internal structure of the R-tree and enumerates the target items in an efficient manner. In the experimental study, it is demonstrated that the proposed algorithm is best suited to the estimation of transition probabilities for relatively small "focused" regions and can facilitate the support of interactive mobility analysis services, which are concentrated on specific areas and need quick response.

Focusing on the maritime domain, Zorbas et al. [115] introduce a machine-learning model using Artificial Neural Networks (ANN), which exploits geospatial time-series surveillance data generated by sea-vessels, to predict future trajectories based on real-time criteria. They model historical patterns of vessels movement in the form of time-series. Their model exploits the past behavior of a vessel in order to infer knowledge about its future position. Their method is implemented by using the MOA toolkit [14][59] and predicts the position of any vessel within the time range of 5 minutes. In that context, the records of a vessel are processed as they arrive in an online fashion and treated as a single trajectory, which directly feeds the forecasting model without taking into account vessels' semantics (i.e., vessel types, geographic area, and other explicit parameters). In their experimental study, the authors evaluate the proposed methodology with respect to the regression accuracy as measured by means of absolute error and execution time. It is shown that their approach supports accurate predictions in the near future, and 30- and 60-minute predictions relatively maintain low mean errors.

In a different domain, that of aviation, Hamed et al. [38] propose a method which predicts the altitude change of an aircraft within a predefined prediction time, i.e., 10 minutes look-ahead, by taking into consideration aircraft's trajectories applied in ground-based systems. Their approach addresses the problem of altitude prediction by utilizing several different methods. They compare the point-mass model with various regression methods (i.e., parametric linear, common parametric non-linear and efficient non-parametric model) to predict intervals, which achieve a desired accuracy ratio of an aircraft's future position. They use radar and meteorological data, collected within a time period of two months, to build a dataset of explanatory and target variables and focus on a single aircraft type (i.e., Airbus A320). Their approach adopts less complex models requiring fewer parameters and thus less trajectories. Principal Component Analysis (PCA) is performed over the data, achieving dimensionality reduction and redundant variables in the computation of the trajectory prediction. They conclude that regression methods compared with the point-mass model achieve more accurate predictions and better results provided that they have used as input historical data from aircrafts' past positions trajectories. In the experimental study, they perform cross-validation by using various methods to support altitude prediction and show that the proposed methodology achieves low prediction error, as measured by means of mean absolute error (MAE) and root mean squared error (RMSE), because it exploits knowledge from past trajectories observations.

### 3.2.2 Techniques based on association rules or frequent / sequential patterns

Apart from traditional classification methods, such as Markov models, the peculiarity of mobility data has also led researchers to use frequent patterns and association rules, as the step upon which they define their prediction model. More specifically, a category of techniques extracts association rules from users' trajectories, which are utilized in a subsequent step to infer a specific user's next movement. The left-hand side (LHS) as well as the right-hand-side (RHS) of these association rules, called *mobility rules* in [107], consist of frequent visited regions. The best rule, according to which the prediction is driven, is selected by using the popular notion of *support*. The formalism of a mobility rule appears below:

$$R : <c_1, c_2, \dots, c_{i-1}> \; \rightarrow \; <c_i, c_{i+1}, \dots, c_k>$$

In this framework, Yavas et al. [107] propose an algorithm for predicting the next inter-cell movement of a mobile user in a PCS network. They address the problem of mining offline mobility data from mobile user trajectories to discover regularities in inter-cell movements, termed as *mobility patterns*, and extract mobility rules from these patterns. Mobility patterns are inferred as the sequence of neighboring cells within the network region traversed by the mobile user, by applying a generalized method of sequential pattern mining over a directed graph. The mobility rules, which match the current trajectory of a mobile user, are used for the online prediction of the user's next movement. In their experimental study, the proposed methodology achieves high precision and recall as the number of the predictions made each time increases.

Yang and Hu [105] propose a model for trajectory patterns and a novel measure representing the importance of a trajectory pattern, to estimate the expected occurrences of a pattern in a set of imprecise trajectories. In particular, they define a novel property, called min-max, and upon it they devise the so-called TrajPattern algorithm. TrajPattern algorithm takes as input parameter *k,* i.e., the number of trajectory patterns that a user aims to find, and produces as output the *k* patterns with the highest Normalized Match (NM). Predictive tasks are supported by using the moving patterns, which are common to a large set of mobile objects which facilitate to obtain the locations of an object in the future. Due to the presence of noise in the trajectories, many similar patterns may be found in the mining process. The concept of *pattern groups* is proposed to compactly represent many similar trajectory patterns via a small number of groups. The TrajPattern algorithm mines the patterns by a *growing* process. It first identifies short patterns with high NM value, and then extends these short patterns to find longer patterns with high NM via the min-max property. With the min-max property, a novel pruning method is derived to reduce the number of candidate patterns, thus the efficiency of the mining algorithm is improved. The TrajPattern algorithm can be used for mining any type of patterns satisfying the min-max property. The experimental study exhibits the effectiveness of the proposed NM model in the FLP task and analyzes the good scalability and sensitivity related with $\delta$, i.e., a small distance unit, of the TrajPattern algorithm.

Verhein and Chawla [97] introduce the *Spatio-Temporal Association Rules (STARs)* to describe the movement of objects from region to region over time. To efficiently deal with the semantics of such data, they define several useful patterns, such as *hot-spots*, which represent the dense regions according to the movement of objects, beiong divided into *stationary* (where many objects tend to remain for long time) and *high traffic* regions (where many objects tend to enter and leave the area during a time period), with the latter being further subdivided into *sources* (high number of leaving objects), *sinks* (high number of entering objects) and *thoroughfares* (both sink and a source). The approach determines which objects are located inside a region, though it does not precisely identify the exact location of an object in that region. The algorithm mines the patterns on a time window basis by performing pruning of search space of STARs based on spatial characteristics and on the observation that only those patterns that have support above a threshold are interesting to a user. This allows for interactive mining as it does not only enable to find current patterns is streaming data but also to capture the evolving nature of the patterns over longer periods of time. In the experimental study, it is demonstrated that the proposed technique has better time performance regarding the rules mining task compared with the brute force algorithm and high precision for large datasets, even when the data are noisy.

Morzy [64] uses an Apriori [1] -like algorithm, called *AprioriTraj*, to discover simple movement rules. The approach exploits the movement rules discovered in a database to unveil frequent trajectories traversed by moving objects and are combined with an object's past trajectory. The movement rules are discovered by iteratively identifying sets of frequent trajectories of length $k$ based on frequent trajectories of length $k$–1. These frequent trajectories are further used as an approximate model per moving object where a fast matching is performed to build its probabilistic model of location. The FLP model is formulated by taking into consideration the movement rules. These rules are then propagated into four methods to retrieve the best matching with respect to a given object trajectory. Each of these methods incorporates a simple, polynomial, logarithmic, and aggregate strategy, respectively, that attempts to find a relative score between long and short movement rules. The approach turns out to be fast as the expensive computations are performed periodically and offline. In the experimental study, preliminary results are shown regarding the performance of the proposed methodology and the efficiency of predicted locations for different rule matching strategies.

In a follow-up work, Morzy [65] proposes the Traj-PrefixSpan algorithm for mining frequent trajectories by modifying the FP-tree index structure [39], originally proposed for efficient frequent pattern mining, in order to achieve fast lookup of trajectories. Traj-PrefixSpan does not allow multiple edges as elements of the sequence, meaning that each element of a sequence is always a single edge. In addition, each sequence is grown only using adjacent edges, and not arbitrary sequence elements. Following the methodology proposed in [39], the Traj-PrefixSpan algorithm consists of three phases: in the first phase the algorithm performs a full scan of the trajectory database to discover all frequent 1-trajectories; in the second phase, each frequent 1-trajectory $Y$ is used to create a $Y$-projected trajectory database, consisting of patterns with prefix $Y$; in the the third phase, the algorithm recursively generates further $Y'$-projected trajectory databases from frequent trajectories $Y'$ found in projections. After frequent trajectories have been found and stored in the (modified) FP-tree, they are used to predict the unknown location of a moving object. In [65], the FLP problem is decomposed into two sub problems, i.e., discovering movement rules with support and confidence greater than user-defined thresholds of *minsup* and *minconf*, respectively, and matching movement rules against the trajectory of a moving object for which the current location is to be determined. An extensive experimental study exhibits the time performance and quality of the predictive tasks. It is also shown that the time needed to mine frequent trajectories remains at manageable level.

Jeung et al. [45] introduce a hybrid prediction model, which exploits moving object's trajectories and motion functions to predict its future location based on a probabilistic approach. They introduce the concept of *trajectory pattern* to model the locations that the user has visited along with the respective time zones. Each specific sequence of visited locations constitutes the premise and the predicted location is the consequence. This correlation between the premise and the consequence comprises the association rules. Also, they present the distant time query, which is the maximum time interval of the prediction length. They discover frequent and similar trajectory patterns by proposing a novel access method, which effectively indexes the constructed association rules and combines them with the object's trajectory. They perform efficient query processing, both for distant and non-distant time queries. Distant time queries (i.e., long-term predictions) are supported by the Backward query processing module whereas non-distant time queries (i.e., short-time predictions) are supported by the Forward query processing module, respectively. These query processing modules are incorporated into the so-called Hybrid Prediction Algorithm (HPA), which predicts future location in a hybrid manner for large spatiotemporal trajectories. In the experimental study, they compare the prediction accuracy and the query response time of the proposed HPM with the RMF approach introduced in [90] (already surveyed in Section **Error! Reference source not found.**). They also investigate the changes of the prediction accuracy by tuning various parameters during the pattern discovery process. It is demonstrated that the HPM exhibits low errors for short-term and long-term predictive tasks and good query response time compared with the RMF approach.

Giannotti et al. [33] propose a sequential pattern mining method, which analyzes the trajectories of moving objects. They introduce trajectory patterns (T-patterns) as concise descriptions of frequent behaviors, in terms of both space (i.e., the regions of space visited during movements) and time (i.e., the duration of movements). T-patterns represent sets of individual trajectories that share the property of visiting the same temporally annotated sequence of places with similar travel times. As a consequence, they introduce the concepts of (i) *Regions of Interest* (RoIs) in the given space, and (ii) the typical travel time of moving objects from region to region. In other words, a T-pattern is a sequence of visited regions, frequently visited in a specified order with similar transition times (with the regions being either set by the user or automatically computed by the method). For instance, $TP_1$ and $TP_2$ below imply that n (m, respectively) trajectories moved from region $A$ to region $B$ at a typical transition time between $t_1$ and $t_2$ (from region $A$ to region C at a typical transition time between $t_1$ and $t_2$ and then to region $B$ at a typical transition time between $t_3$ and $t_4$, respectively).

$TP_1$: <0, A> <( $t_1$, $t_2$), B> (supp: n)
$TP_2$: <0, A> <( $t_1$, $t_2$), C> <( $t_3$, $t_4$), B> (supp: m)

This methodology is not used per se to support FLP tasks, however it introduces an efficient approach towards the trajectory pattern mining problem. Therefore, in a follow-up work, Monreale et al. [62] propose the so-called *WhereNext* method, which is a sequential pattern mining approach building upon T-patterns. Precisely, the *WhereNext* methodology first extracts a set of T-patterns and then constructs a T-pattern tree, which is a prefix-tree built upon the frequently visited sequence of regions of the discovered T-patterns. In this process, a T-pattern is considered as a prefix of another T-pattern if and only if the cardinality (i.e., the size / length of the sequence) of the first is equal to or smaller than that of the second and the sequence of all regions of the first is a sub-sequence of that of the second. Interestingly, in order to produce a compact T-pattern tree and in order to account also the transition times and the support values of the original T-patterns, a unification process takes place during tree construction. In this connection, each path of the tree is a valid T-pattern; an example of T-pattern tree is illustrated in Fig. 8.

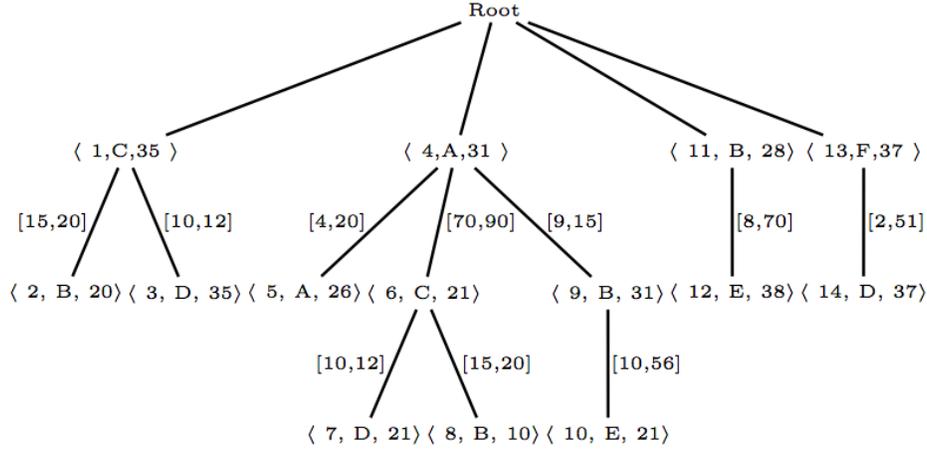

Fig. 8: An example of T-pattern tree. [Figure source: [62]]

After the T-pattern tree has been built, a given trajectory (for which we aim to make a prediction) is used to find the best matching score with respect to all paths (i.e., T-patterns) that fit it. This fitness score takes into account whether the trajectory spatially intersects a node (i.e., a frequent region) of a path of the tree, or, if it does not, it accounts the spatiotemporal distance of the trajectory with respect to the tree node. In the end, the predicted future location of the trajectory is the region that corresponds to the final node of the best path. In the experimental study presented in [62], the accuracy of the proposed model with respect to the prediction rate in different contexts is exhibited. It is demonstrated that the proposed methodology gives accurate prediction for a reasonable set of trajectories and enables users to tune the algorithm by using a set of thresholds. It is noteworthy that, among the papers surveyed, this work is chronologically the first to use a real dataset (although not available for public use), in particular, a set of trajectories from 17,000 GPS-equipped cars moving in the city of Milan, Italy, over a week.

Towards a slightly different objective, Zheng et al. [112] consider a user's travel experience and the locations that he/she has sequentially visited to propose an approach which mines the correlation between locations from many users' location histories. These correlations indicate the relationship between locations and human behavior in the space, and enable many valuable services, such as sales promotion and location recommendation. They propose a personalized location recommendation system by using the location correlation. This approach facilitates the prediction of an individual's preference in various locations by using their location history as well as those of other people. It involves a method to uniformly model each individual's location history and the design of a model, which infers each user's travel experience in a given geo-region. The algorithm, which learns the correlation between locations, considers users' travel experiences and the sequence of the locations in a user's trip. The recommendation system is deployed by using a Collaborative Filtering (CF) algorithm and exhibits high efficiency compared with the Pearson correlation-based CF model [73]. In their experimental study, also based on a real dataset (a population of 112 users recording their outdoor activities for a period of 20 months), the authors present the effectiveness of their method in terms of comparing with different methods in the personalized location recommendation and efficiency in conducting the predictive tasks. It is shown that the proposed methodology achieves good results with respect to the baseline methods.

From the same group, Ye at al. [108] propose the novel notion of *individual life pattern*, which captures individual's general life style and regularity. To formally describe which kind of life regularity can be discovered from location history, the so-called *life pattern normal form* (LPnormal form) is introduced, which focuses on significant places of individual life and considers diverse properties to combine the significant places. Their approach encapsulates LPnormal form to the LP-Mine framework to effectively retrieve life patterns from raw individual GPS data. LP-Mine is comprised of two phases: the modeling phase and the mining phase. The modeling phase preprocesses GPS data into an available format which it then constitutes the input of the mining phase, where it is accomplished in two steps, i.e., the detection and the density-based clustering of stay points. The mining phase applies separate methodologies for different types of patterns, including temporal sampling and partition, extracting temporal, non-temporal and conditional life patterns, frequent itemset and sequence mining and corset discovery, to discover different types of pattern. The mining phase facilitates to predict future life trends based on patterns harvested from historical life patterns. In their experimental study, the authors demonstrate the efficiency of the patterns prediction module for different support thresholds regarding sequential and non-sequential life patterns.

Gomes et al. [34] propose NextLocation, a personalized mobile data mining framework that uses spatial, temporal and other contextual data (accelerometer, Bluetooth, and call/sms log) to address the FLP problem. Predicting the next location is reduced to a classification problem, and the methodology consists of three steps, i.e., data preprocessing, anytime model and accuracy estimator. Data preprocessing involves the transformation of data for next place prediction; the spatial data from a visit is enriched with other contextual information. The preprocessing component only requires keeping a short-term window of data. The anytime model integrates new information as it is available (such as new visits) and is able to predict the next location. Hoeffding Trees, available in the MOA toolkit [14][59], are used as the base learner to support a drift detection on the data. The J48 classification algorithm for decision tree induction, available in WEKA platform [101], can incrementally learn from the data, hence it is considered the most appropriate to adapt the model according to the changes in the user mobility patterns. The accuracy estimator compares the output from the anytime prediction model with the actual destination and allows to keep an estimate of data prediction accuracy. NextLocation achieves reduced communication overheads in terms of bandwidth as well as battery drain since the processing is made locally avoiding expensive wireless data transfer. It also enables an alternative business model where advertisement providers can push content that is relevant to a certain location and the user will only receive it when is about to visit it. In the experimental study, using real data from the Nokia Mobile Data Challenge (MDC) [51], the proposed approach is evaluated across different users and turns out to achieve high percentage of predictive accuracy for regular moving objects.

MyWay by Trasarti et al. [96] addresses the FLP problem by exploiting users' systematic mobility in the form of individual systematic behaviors modeled by mobility profiles built upon users' history; based on these profiles, MyWay is able to forecast the exact future position of mobile users at specific time instants. The prediction method consists of three prediction strategies, i.e., an individual strategy regarding the regularity of a single user (systematic behavior of the user), a collective strategy that takes advantage of the individual systematic behaviors of users, and a hybrid strategy that combines the two above levels of knowledge, applying the collective strategy when the individual one fails. Towards this direction, their approach adopts the concept of Personal Mobility Data Store (PMDS), where each user stores the information regarding her movement data separately. In order to obtain better results in terms of profile quality and efficiently deal with trajectories having different sampling rate, they define a new distance function called Interpolated Route Distance (IRD) as well as a variation of it, called Constrained IRD (CIRD), by taking into account the aspects of interpolation and symmetricity among the trajectory samples. In their experimental study, based on a large real dataset (9.8M car travels performed by about 159K vehicles[4]), the authors present the effectiveness of their method compared to the prediction performances of individual and global competitors, considering also the parameters regarding to the disclosure of data, the computational cost and the system's capability to deal with a real big data context. Therefore, they highlight that the synergy between the individual and collective knowledge is the key for a better prediction.

Millefiori, Vivone, et. al. [126][127] present a novel method for predicting long-term target states employing an Ornstein-Uhlenbeck (OU) stochastic process, providing target state equation revisions that leads to orders of magnitude lower uncertainties in the time scaling than under the nearly constant velocity (NCV) assumption. Application of the proposed model on a significant portion of real-world maritime traffic provides promising results.

---

[4] To our knowledge, this dataset is the largest among the ones that have been used in the literature so far for FLP/TP tasks; unfortunately, the dataset is not available for public use do to the country privacy law, as declared in [96].

### 3.3 Semantic-aware methods

The semantic-aware prediction of moving entities, which is realized by harvesting enriched spatiotemporal data with contextual and other behavioral characteristics, has attracted a lot of attention in the recent years. However, most of the techniques only focus on the geographic characteristics of the generated trajectories.

Ying et al. [110] are the first who exploit both geographic and semantic features of trajectories. Their approach is based on a novel cluster-based prediction method, which estimates a mobile user's future location by exploiting frequent patterns in similar users' behavioral activities. Users' similarity is determined by grouping the common behavior of users in semantic trajectories. The outcome of this approach is the SemanPredict framework, which consists of two modules, i.e., the (i) offline and (ii) online mining module. The framework takes into consideration GPS trajectories as well as cell trajectories, which are trajectories that are lying in wider geographic areas. The authors also introduce the concept of *stay location*. Stay location represents the areas that the users have stayed for a specific time interval. The offline mining module, serving as a preprocessing stage, transforms individual user's semantic trajectories into a labeled sequence of stay locations and the result is stored in a tree-based structure, titled as Semantic Trajectory Pattern tree (STP-tree); an example of STP-tree is illustrated in Fig. 9. Then, the method uses this encoded user activity in order to determine and cluster similar sequences of stay locations among different users. The online mining module, supporting the prediction method, exploits both a single user's profile and the profile of his/her group produced by the clustering method. This enables to compute the best matching score between the candidate paths and efficiently predict his/her next stay location. In the experimental study, the prediction accuracy is measured by means of precision, recall and F-measure, and is demonstrated that the proposed approach has good performance under various conditions, i.e., parameters setting, impact of the semantic clustering, prediction strategies and execution time.

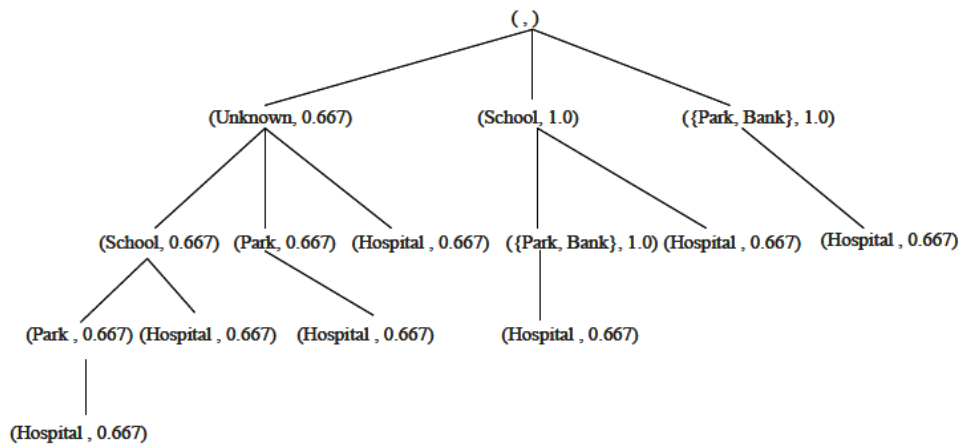

Fig. 9: An example of STP-tree. [Figure source: [110]]

Wu et al. [102] introduce a novel location prediction method, titled as spatial-temporal-semantic long short-term memory (STS-LSTM) algorithm, which exploits the spatial, temporal, and semantic information of the raw trajectory. Their method approaches the location prediction as a classification problem, incorporating the concept of discrete locations. The proposed spatial-temporal-semantic feature extraction algorithm aims to discretize the trajectories into long location sequences, that are exploitable during the prediction process. More specifically, the STS algorithm transforms the trajectory into location sequences with fixed code (i.e., fixed and discrete road IDs), taking the points along the road into account, maintaining this way as much information as possible. The trajectory discretization is being accomplished by adding both spatial and temporal factors into the model. During the temporal mapping process, each trajectory is divided into segments, and then track points are allocated to *timebins*, where for each timebin a representative point is selected. In order to map the representative points to fixed reference points on the road, a new geographical feature matching method is introduced, exploiting OpenStreetMap (OSM) data and integrating the semantic tags of each road in the network. In that context, after the spatial-temporal-semantic feature extraction, the trajectory is transformed into a location sequence. The proposed long short-term memory (LSTM) -based neural network model, is based on a conveyor belt-like structure, and can take advantage of location sequences over a long period of time, achieving stable and higher prediction accuracy compared with traditional feature extraction and model building methods (e.g. HMM or RNN). As this method becomes suitable for exploiting the points along the road and can handle long location sequences, the accuracy of the algorithm is highly dependent on the quality of the trajectory, while the model cannot accept more features and dimensions.

### 3.4 Summary

In this section, we provided a survey of over twenty (precisely, twenty-four) FLP techniques, actually falling in one of three categories: (i) those performing short-term predictions based on the motion vectors of moving objects (typically, points), which are appropriately indexed in spatial access methods (R-trees, etc.), (ii) those predicting future locations by exploiting patterns (classification trees, association rules, etc.) over the past movements of objects, and (iii) those applying core machine learning approaches (SVR, NN, etc.), where models are trained by pat movements. Table I summarizes the most representative works, also giving notes on the datasets used for the experimental evaluation of the corresponding proposals.

Summarizing, we provide the following remarks (and hints for practitioners) about the surveyed FLP techniques:

– Vector-based approaches, inspired by the spatial database management domain, aim to model current locations (and perhaps short history) of objects as motion functions in order to be able to predict future locations by some kind of extrapolation. This is more or less the concept and the state-of-the-art in this category of methods includes, at least, TPR*-tree [91], B$^x$-tree [44], and STP-tree [90].

– Pattern-based approaches, inspired by the spatial data mining domain, exploit on data mining patterns (classification models, frequent / sequential patterns etc.) that are built upon the history of movements. Under this prism, the prediction task is an instantiation of the pattern that most fits to the short history of the object's movement (the 'tail' of its trajectory so far). State-of-the-art methods in this area include, at least, NextLocation [34], MyWay [96], Ornstein-Uhlenbeck (OU) stochastic process [126][127].

— Semantic-aware approaches involve semantics extracted by the surrounding environment (whether e.g. a stay is at home or at a park), build patterns on top of this knowledge, and apply them for prediction of the next location(s). Since this is the most recent approach there are only a few related works, among them we may consider SemanPredict [110] and STS-LSTM [102] as state-of-the-art.

Table I: Surveyed FLP techniques.

| Ref. in this paper | Technique | Input | Output | Dataset(s) used for the empirical evaluation |
|---|---|---|---|---|
| cf. section 3.1 | PMR Quadtree by Tayeb et al. [93] | Motion vectors of moving objects (points) | Near-future locations (short-term) | Synthetic dataset simulating 50K moving points; (ad-hoc) movement states are simple linear functions |
| | Dual space-time representation by Kollios et al. [49] | Motion vectors of moving objects (points) | Near-future locations (short-term) | Synthetic dataset simulating 100K − 500K moving points; (ad-hoc) movement states are simple linear functions |
| | SUTR, QTR+QR by Xu & Wolfson [104] | Motion vectors of moving objects (points) | Near-future locations (short-term) | Synthetic dataset simulating 5K trajectories of moving points; (ad-hoc) movement states are simple linear functions |
| | TPR-tree by Saltenis et al. [79] | Motion vectors of moving objects (points) | Near-future locations (short-term) | Synthetic dataset simulating 100K (scaled up to 900K) points moving to 20 (scaled up to 100) destinations |
| | TPR*-tree by Tao et al. [91] | Motion vectors of moving objects (points) | Near-future locations (short-term) | Synthetic dataset simulating 100K points moving to 5K destinations |
| | STP-tree / RMF by Tao et al. [90] | Motion vectors of moving objects (points) | Near-future locations (short-term) | Synthetic datasets simulating (a) typical movement curves, such as polynomial, sinusoid, etc. (RMF effectiveness); (b) 10K points moving to various destinations (STR-tree efficiency) |
| | B$^x$-tree by Jensen et al. [44] | Motion vectors of moving objects (points) | Near-future locations (short-term) | Synthetic datasets simulating 500K (scaled up to 1M) points moving (a) randomly; (b) according to a road network movement generator [79] |
| cf. Section 3.2.1 | LeZi-Update by Bhattacharya & Das [12] | Trajectories, in a granularity of cell | Next cell | Synthetic datasets simulating cell-to-cell movement (from a population of 8 cells) |
| | CSP-based algorithm for mobility statistics by Ishikawa et al. [42] | Trajectories, as sequences of points, over a cell partitioning | (cell-to-cell) Transition sequence enumeration | Synthetic trajectory datasets using Brinkhoff generator [16]: ~100 objects are alive at any time on the network, ~125K points in total, varying (up to 30x30) cell partitioning of space |
| | ANN models by Zorbas et al. [115] | Trajectories, as sequences of points (vessels only) | Next locations, ~5 min. look-ahead | Real dataset, consisting of 2874 vessel routes (about 1.3M points − AIS signals) operating across the Aegean Sea for a duration of 23 days; marinetraffic.com property, N/A for public use. |

| Ref. in this paper | Technique | Input | Output | Dataset(s) used for the empirical evaluation |
|---|---|---|---|---|
| | Regression models by Hamed et al. [38] | Trajectories, as sequences of points (aircrafts only), weather, etc. | Next locations (altitude only), ~10 min. look-ahead | Real dataset, consisting of 1500 Airbus A320 aircraft flights departing from Paris airports (Paris Orly and Paris Roissy- Charles de Gaulle), 10 min takeoff phase, varying sampling, altitude info only (altitude is in Mode C (ft x100); Paris Air-Traffic Control Center property, N/A for public use. |
| cf. Section 3.2.2 | Mobility rules by Yavas et al. [107] | Trajectories, in a granularity of cell | Next cell | Synthetic dataset consisting of 10K (ad-hoc) trajectories; fixed (15x5 hexagonal) cell partitioning of space |
| | AprioriTraj by Morzy [64] | Trajectories, as sequences of points, over a cell partitioning | Next cell(s) | Synthetic dataset consisting of 300 trajectories (scaled up to 5K) using Brinkhoff generator [16]; varying grid size |
| | Traj-PrefixSpan by Morzy [65] | Trajectories, as sequences of points, over a cell partitioning | Next cell(s) | Synthetic dataset consisting of 1K trajectories (scaled up to 10K) using Brinkhoff generator [16]; varying grid size |
| | STAR by Verhein & Chawla [97] | Trajectories, as sequences of points | Next region(s) | Synthetic dataset consisting of 10K moving points; varying number of regions |
| | TrajPattern by Yang & Hu [105] | Trajectories, as sequences of points | Next location(s) | Synthetic dataset consisting of a small number of trajectories (undefined in [105]), based on ZebraNet dataset [53] |
| | HPA by Jeung et al. [45] | Trajectories, as sequences of points | Next location(s) | Synthetic datasets using a modification of the periodic data generator [54] and simulating the movement of cars, bikes, cows, and planes, based on a small number of respective real datasets; N/A for public use. |
| | T-pattern tree by Monreale et al. [62] | Trajectories, as sequences of points | Next location(s) | Real dataset, consisting of the trajectories of 17K GPS-equipped cars moving in the city of Milan, Italy, over a week; N/A for public use. |
| | Location correlation by Zheng et al. [112] | Trajectories, as sequences of points | Location recommendations | Real dataset, consisting of the trajectories of 112 GPS-equipped users moving in Beijing and other cities of China for a period of 20 months, resulting in a total of ~9.5M points; the specific dataset is N/A in the original paper, however, it is part of GeoLife trajectory dataset, which is available at [113] |

| Ref. in this paper | Technique | Input | Output | Dataset(s) used for the empirical evaluation |
|---|---|---|---|---|
| | Individual life patterns by Ye et al. [108] | Trajectories, as sequences of points | Future life trends | Real dataset, consisting of the trajectories (of several · the specific number is undefined in the paper · GPS-equipped users moving in China and elsewhere, covering a total of over 50,000 km; the specific dataset is N/A in the original paper, however, it has some overlap with the GeoLife trajectory dataset, which is available at [113] |
| | NextLocation by Gomes et al. [34] | Trajectories, as sequences of point, along with other related data (accelerometer, Bluetooth, call / SMS log) | Next location(s) | Real dataset, consisting of the trajectories (and other related information) of 200 smartphone-equipped users; the dataset was released for the purposes of Nokia MDC [51] and is available at [60] |
| | MyWay by Trasarti et al. [96] | Trajectories, as sequences of points | Next location(s) | Real dataset, consisting of a total of 9.8M trajectories from 159K GPS-equipped cars moving in Tuscany, Italy, for a period of one month (May 2011); N/A for public use. |
| cf. Section 3.3 | SemanPredict by [110] | Trajectories, as sequences of cells | Next location(s) | MIT reality mining dataset [24]: a mobile phone dataset collected by MIT Media Lab from 2004 to 2005, recording the activities of 106 mobile users (over 500K hours, in total); available for public use at [58]. |
| | STS-LSTM [102] | Trajectories, as sequences of points | Next location(s) | Two real datasets: (a) courier dataset in Beijing (66 days, 198 trajectories), N/A for public use; (b) one bus route in New York from MTA bus dataset (90 days, 149 trajectories), available for public use at [66]. |
| | Ornstein-Uhlenbeck (OU) [126][127] | Probabilistic motion equations | Next location(s) | HFSW radar data (WERA) from Palmaria, Italy (2009); JPDA Tracker for HFSW radar data; SAR data acquired by Sentinel-1A (ESA); AIS data in combination with the above. |

## 4. TRAJECTORY PREDICTION

Recalling Problem 1 (FLP) and Problem 2 (TP) definitions presented in Section 2, it turns out that, in fact, an FLP method could be transformed to address the TP problem, given a specific granularity upon which the same method can be applied iteratively. The main difference with respect to "pure" TP methods is that in the FLP-aiming-to-solve-TP case the prediction errors are accumulated with each step (e.g. via multi-step Linear Regression), thus making the predicted points increasingly error-prone. In contrast, pure TP methods aim to forecast the complete trajectory from the start, thus making each predicted point equally error-prone. Not surprisingly, the vast majority of methods are domain-specific (with most of them in the aviation domain), in order to take advantage of the properties of the moving objects under consideration.

### 4.1  Stochastic approaches

This category of trajectory prediction techniques enables the handling of sources with uncertainty in a wide range of alternatives. Under this perspective, stochastic airspace models enable the reflection of the unexpected impact of both known and unknown factors regarding the forecasting trajectory. The uncertainty associated with trajectory predictions, particularly the knowledge of the sources of uncertainty, facilitates the prediction task and ensures more accurate estimation results.

Ayhan and Samet [6] introduce a novel stochastic approach to the aircraft TP problem, which exploits aircraft trajectories modeled in space and time by using a set of spatio-temporal data cubes. They represent the airspace in 4-dimensional joint data cubes (i.e., latitude, longitude, altitude, and time) consisting of aircraft's motion track in space-time and enriched with weather conditions. They use the Viterbi algorithm [98] to compute the most likely sequence of states derived by a Hidden Markov Model (HMM) [77], which has been trained over historical surveillance and weather conditions data. The algorithm computes the optimal state sequence in the maximum likelihood sense, which is the one that is best aligned with the observation sequence of the aircraft trajectory; the gist of their methodology is illustrated in Fig. 10. In their experimental study, it is demonstrated that the proposed methodology predicts aircraft trajectories efficiently by comparing the prediction results with the ground truth trajectories of the testing set, aligned to this grid of spatio-temporal cubes. They also measure, by means of mean and standard deviation values, the horizontal, vertical, along-track, and cross-track error and present that the error is reasonably low, i.e., it resides within the boundaries of the highest spatial resolution of the grid.

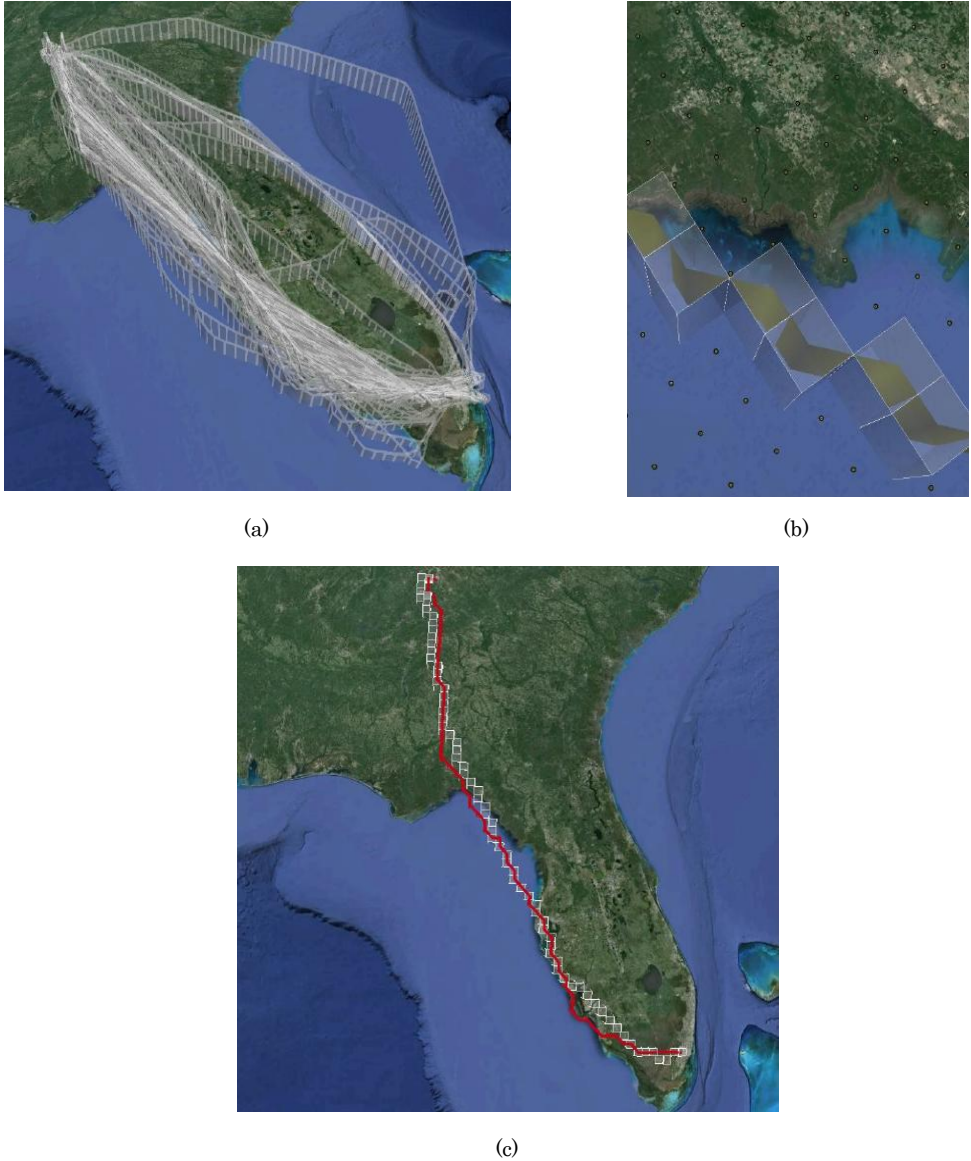

Fig. 10: TP with HMM: (a) historical trajectory data of flights; (b) discretization of the spatio-temporal space into a state-transitions grid; (c) TP with the trained HMM [Figure source: [6]].

One very important aspect in state-based TP is the segmentation of an aircraft trajectory into distinct phases or *segments*. Gong and McNally [35] propose a methodology for automated TP analysis, specifically for splitting the process in separated stages according to the flight phases. The purpose is to identify flights, as described by actual radar tracks, which show unpredictable modifications of their aircraft intent and can be considered outliers. This segmentation process is of high interest when preparing such a dataset for other machine learning algorithms in the context of TP.

The concept of stochastic modeling for TP is applied in a different way by Ramos [78]. Safety constraints in ATM dictate that it is more reasonable to predict spatio-temporal intervals rather than precise aircraft positions, as in this manner it is easier to have proper probability estimations and confidence levels with respect to various uncertainty factors in the input. As in HMM, the use of intervals essentially transforms the continuous spatio-temporal TP problem into a discrete state-transition model. In this work, a kinematic stochastic model was used, associated with a probabilistic performance model that captures the variability associated with the execution of a flight phase, combined with the Monte Carlo method for reproducing the possible trajectory paths. The training data were obtained from Automatic Dependent Surveillance Broadcast (ADS-B) system [25]. ADS-B broadcasts the aircraft's id, location and operational information to other aircraft and ground stations in the vicinity. The ADS-B receptors provide the location data, which are enriched with the type and flight phase of the aircraft, while the flight phases were identified using the Viterbi algorithm.

## 4.2  Regression and clustering -based approaches

As expected, Neural Networks (NN) have been proposed in various works as the core regression model for the task of TP.

Le Fablec et al. [52] introduce NNs for the specific problem of predicting an aircraft trajectory in the vertical plane, i.e., its altitude profile with the time. Two separate configurations are considered: (a) the case of *strategic* prediction, where the aircraft has not taken off yet; and (b) the case of *tactical* prediction, where the already flown aircraft states are used to improve the prediction. The proposed NN is a feed-forward model with one hidden layer, using the aircraft type and the difference between the Requested Flight Level (RFL), which defines the planned cruising altitude, and the actual altitude, as its main inputs.

Cheng et al. [18] employ a data mining statistical approach on the radar tracks of aircrafts to infer the future air traffic flows using NNs. More specifically, a training dataset is produced using the radar tracks grouped in seven 'weekday' categories and the Estimated Time of Arrival (ETA) at designated fixes and airports as output. The NNs used feed-forward architecture with back-propagation and one hidden layer of 5-10 neurons, with one such model trained separately for each weekday (category).

In a similar approach, Hong and Lee [41] introduce a TP method for vectored area navigation arrivals. This practically translates to predicting arrival times by leveraging probabilistic information about the trajectory management patterns from Air Traffic Control (ATCO) to ensure safe and efficient operations. The method considers different aircraft types and approach patterns, the likelihoods of which are estimated by analyzing previous radar tracks, predicting the ETA to the runway considering the time at entry fix. The major patterns of vectored trajectories are identified by clustering the recorded radar tracks for the airspace of interest. The clusters are built upon the computation of the relative Euclidean distance of one trajectory from the other. Corrections of time misalignments among trajectories are addressed via Dynamic Time Warping (DTW) [47][70][68]. Finally, multiple linear regression models for travel time are formulated for each of those identified patterns. Then, for a new inbound aircraft of specific type and entry fix, the trained model chooses the most suitable approach pattern and ETA.

De Leege et al. [21] also address the specific TP task of predicting arrival routes and times via a slightly different approach. A Generalized Linear Model (GLM) is employed for the merging of air traffic following fixed arrival routes, together with meteorological data and two aircraft types, according to the availability of ADS-B data. Stepwise linear regression was applied to determine which regressors to include in the GLM, adding or removing regressors based on their statistical significance in explaining the output variable. In this case, only arrival time predictions for aircraft following fixed arrival routes were investigated.

Yang et al. [106] investigate the same TP task of aircraft intent in the terminal phase using a two-phase online trajectory clustering. The first stage identifies the associated intent model, while the second one calculates the specific intent based on the knowledge of the referred model. The intent modeling is essentially an online trajectory clustering problem, where the real-time routes are represented by dynamically updated cluster centroids extracted from radar tracks without flight plan correlations. Subsequently, the intent identification is a probabilistic scheme integrating multiple flight attributes, including call sign, destination airport, aircraft type, heading angle, etc.

Kun and Wei [50] propose a two-fold 4-dimensional model for both strategic and tactical TP focused on flying time versus positions and altitudes. The first prediction is performed by using a multiple regression that relates the influences of traffic flow and wind conditions. The second prediction requires the normalization and grouping of flying positions and altitudes from different trajectories (radar tracks) to the same time interval.

Tastambekov et al. [92] illustrate a somewhat different regression-based approach for short/mid-term TP, i.e., estimation of where an aircraft will be located over a 10-30 minutes time frame. This task is relevant to several ATCO operations such as Conflict Detection (CD) and Conflict Resolution (CR). This approach is based on local linear functional regression that considers data preprocessing, localizing and solving linear regression using wavelet decomposition, considering a time-window between 10 and 30 minutes. The learning process is designed in two separate stages: (a) localization of data using k nearest neighbors (kNN) for selecting the relevant trajectories; and (b) solving of regression using wavelet decomposition in Sobolev space. The results show that this method exhibits a high level of robustness, although it does not consider the effects of the weather conditions (e.g. wind) in the prediction model.

Song et al. [85] propose a combination of clustering and Kalman filters for TP. Specifically, historical radar tracks are processed to derive typical aircraft trajectories by applying DBSCAN clustering algorithm [27]. Subsequently, the representative trajectory is used to feed a hybrid predictor that instantiates an Interacting Multiple Model (IMM) Kalman filter [40][76]. Using the typical trajectory ensures that the associated flight intent represents the intended trajectory better and, hence, limit the errors in long-term TP.

### 4.3  Collaborative approaches

Stochastic, regression-based and clustering-based approaches are by far the most popular families of methods for the TP task. However, other works approach the problem from a different perspective than that of the single-object tracking. Moreover, there is an increasing interest in integrating TP with other safety-critical task, such as Conflict Resolution (CR) and strategic planning of flights in the context of Air Traffic Management (ATM).

Xiangmin et al. [103] provide an approach for flight Conflict Avoidance (CA) based on a memetic algorithm. The current approaches for CA are often focused on a short-term elimination of conflicts via local adjustment; hence, they cannot always provide a reliable global solution. In contrast, long-term conflict avoidance approaches provide solutions via strategically planning traffic flow from a global point view, a process, which entails TP solutions for the air traffic under consideration. In order to address this large-scale combinatorial constrained optimization problem and avoid local minima, their work presents an effective approach based on a memetic algorithm (called MA) for global search capabilities, as well as a fast genetic algorithm (called GA) as the global optimization method.

Matsuno et al. [57] address the similar problem of CR in ATM by employing a stochastic optimal control approach under wind uncertainty. Their method is used for determining 3-dimensional conflict-free aircraft trajectories including wind parameters by employing (a) a spatially correlated wind model to describe the wind uncertainty and (b) a probabilistic Conflict Detection (CD) algorithm using the generalized polynomial chaos method. The proposed approach is presented as a promising choice w.r.t. a reactive algorithm for CA of multiple realistic Unmanned Aerial Vehicles (UAVs). This task is also addressed by Manathara and Ghose [55], Albarker and Rahim [2], using other methods like agent-based collaborative CA.

Baek and Bang [8] explore the ADS-B data for TP and CD purposes in ATM. The proposed method explores the usage of ADS-B data for producing multiple-model based TP with accurate predictions of conflict probability and conflict zones (blocks) at a future forecast time frame.

## 4.4 Other related approaches

A somewhat different approach in the formulation of the CR-related problems is via the complex network analysis, particularly the detection and minimization of potential conflicts, either in the context of pre-flight strategic planning or of the online tracking and predictive analytics on concurrent flights. For instance, Chen et al. [17] illustrate a framework of Conflict Detection and Resolution (CD&R) in air traffic via complex network analysis by employing collaborative control. In particular, they examine the applicability of alternative automated CD&R concepts for en route ATCO by taking into consideration the underlying network architecture. The proposed design framework highlights several issues regarding the vulnerabilities and the potential for improvement of the current air traffic control system. Performance measures, i.e., accuracy, adaptability, availability, fault tolerance, reliability and dependability, are used to evaluate the different CD&R concepts under the prism of both analytical and simulation methods. Among other theoretical and practical issues, emergent insights from this analysis demonstrate the correlation between network architecture and number of conflicts to resolve (e.g., centralized networks (CNs) [94] and Bose–Einstein condensation networks (BECNs) [13] eliminate the number of potentially occurring conflicts). In addition, the capability of each CD&R concept regarding the traffic density is mainly based on the network connectivity and vulnerability; e.g., scale-free networks (SFNs) [9] and Random networks (RNs) [84][26] are capable of supporting higher densities and are proven more resistant to targeted attacks, compared to the centralized schemes.

Di Cicio et al. [23] explore solutions for the problem of detecting flight trajectory anomalies and predicting diversions in freight transportation. When an airplane diverts, logistics providers must promptly adapt their transportation plans in order to ensure proper delivery despite such an unexpected event. However, different parties in a logistics chain do not exchange real-time information related to flights. Hence, the detection of diversions needs to be calculated proactively using publicly available data, without knowing the exact planned trajectory of a flight. This work addresses this challenge via a prediction model that includes the aircraft position, velocity and intended destination, and the algorithm essentially functions as an anomaly detector.

Yepes et al. [109] consider a hybrid estimation algorithm called the residual-mean interacting (IMM) to predict future aircraft states and flight modes, exploiting information from ATM regulations, flight plans, pilot intent and environmental conditions. The intent inference process is posed as a discrete optimization problem whose cost function uses both spatial and temporal information. Using ADS-B data, the algorithm estimates the likelihood of possible flight modes and selects the most probable one. The trajectory is then determined by a sequence of flight modes that represent the solvable motion problems to be integrated, in order to obtain the corresponding trajectory.

Swierstra and Green [86] provide a system engineering approach for investigating important design issues and tradeoffs, such as the balance between TP accuracy and computational speed. Key aspects of a common TP module are presented, including an approach to dynamically adapt the performance to support a variety of TP applications. The characteristics of different aircraft performance models, the flight path integration logic and software implementation issues are also discussed.

## 4.5 Semantic-aware techniques

Advances in aviation domain are on the road of establishing semantic-aware TP capabilities as an essential building block for efficient and safe operations, with applications that range from en route aircrafts to terminal approach advisory. Precise STP also enables better estimations for departure and arrival times and, hence, more robust scheduling and logistics, especially in the congestion points (airports, major waypoints, etc.).

Regarding en route climb TP, one of the major aspects of decision support tools for ATM, Coppenbarger [20] discusses the exploitation of real-time aircraft data, such as aircraft state, aircraft performance, pilot intent and atmospheric data for improving ground-based STP. A similar approach has been the focus of innovative designs of systems, such as the development of the Center-TRACON Automation System (CTAS) at NASA Ames (Slattery [83]). In this latter work, the general concept of CTAS is investigated in the context of en route climb STP accuracy using available flight-planning data; a typical model for climb-based TP is illustrated in Fig. 11. The results confirm the significant impact of declared takeoff weight, speed profile and thrust calibration data on CTAS climb TP accuracy.

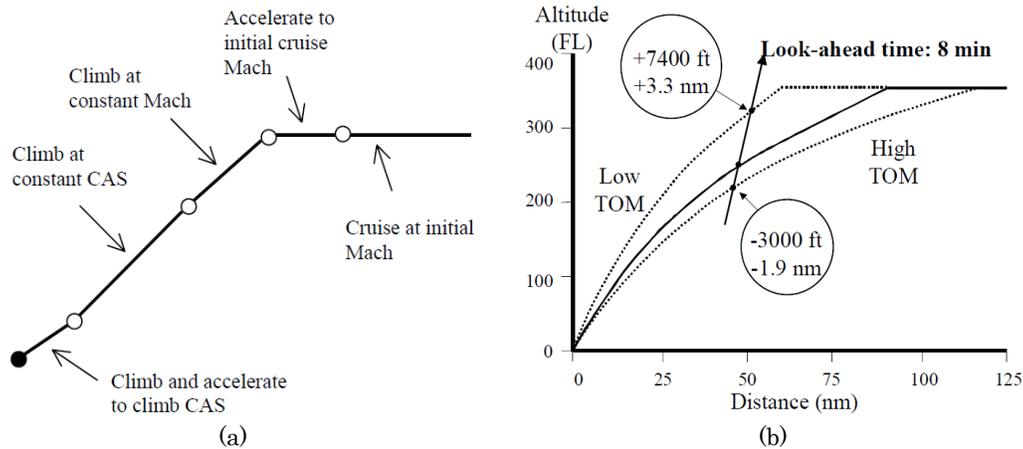

Fig. 11: Typical model for climb-phase TP: (a) Constant Airspeed (CAS) and Constant Mach (CM) stages to departure/cruising altitude; (b) prediction margins for ground-based TP with low/high Takeoff Mass (TOM) and actual climb path (solid line) [Figure sources: [83]]

The problem of climb TP is also discussed by Thipphavong et al. [95], as it constitutes a very important challenge in ATM. Aircraft climb trajectories are difficult to predict because small variations or estimation errors in the model parameters, especially aircraft weight and wind, result in large prediction errors even in the short-term. In this work, an algorithm that dynamically adjusts modeled aircraft weight is developed, exploiting the observed track data to improve the TP accuracy for climbing flights. Real-time evaluation with actual air traffic data shows a significant improvement on the prediction of the trajectory altitude, as well as the time to reach the top-of-climb.

Hadjaz et al. [37] present an approach that is based on a hybrid system to numerically simulate the climbing phase of an aircraft. The Covariance Matrix Adaptation Evolution Strategy (CMA-ES) optimization algorithm is used to adapt five selected parameters, in order to improve the accuracy of the analytical model. Experiments with the proposed model included (a) estimation of errors along time for a one-time TP at the take-off phase of the flight, with respect to the default values of the theoretical Base of Aircraft Data (BADA) point-mass model, and (b) application to on-line TP, in which the prediction is continuously updated based on the current aircraft position. The resulting hybrid TP method shows statistically significant improvements over prediction accuracy compared to the default model.

Recently, Ayhan and Samet [7] investigated the applicability of the HMM for the climb after takeoff. Moreover, they addressed the problem of incorporating weather conditions in their model, as they represent a major factor of uncertainty in all TP-related applications. A stochastic approach such as the HMM can address the STP problem by taking environmental uncertainties into account and training a model using historical trajectory data along with weather observations. However, every aircraft trajectory is associated with different weather conditions on different dates and times. Hence, it is imperative that these sets of parameters are unified in a systematic way, so that they can be incorporated into the HMM as observations (emissions). In this work, a time series clustering algorithm is employed to generate an optimal sequence of weather observations; specifically, the k-Nearest Neighbors (k-NN) algorithm is implemented with Dynamic Time Warping (DTW) Euclidean distance. The results show robust performance and high prediction accuracy, proving that HMM can be applied equally well for single-phase prediction, as well as complete-flight prediction in the general TP context, as described earlier [6].

## 4.6 Summary

In this section, we provided a survey of over twenty (precisely, twenty-five) TP techniques and variants, which fall into five main categories. Most state-of-the-art methods can be labeled as stochastic, regression-based, clustering-based or some hybrid combination of these basic approaches. Especially for the aviation domain, other methods address the TP problem indirectly, under the collaborative scope of conflict detection and resolution (CD&R) or collision avoidance (CA). The climb phase TP is of particular importance and it requires the embedding of other external parameters, including weather, aircraft properties, etc., with methods typically referred to as semantic-aware TP due to their inherent increased dimensionality (enriched spatio-temporal input space). Table II summarizes the most representative works and provides additional notes with respect to the datasets used in the corresponding experimental evaluation in each one.

Summarizing, we provide the following remarks (and hints for practitioners) about the surveyed TP techniques:

- Stochastic approaches formalize the TP task via designing proper probabilistic models based on historical trajectory data and then providing predictions in the maximum-likelihood sense, typically by some form of HMM or similar formulation; current state-of-the-art includes, at least, HMM-based [6], segmented/phased trajectories [35] and confidence interval-based [78].
- Regression-based methods provide abstract functional mapping between the input and the output (typically the spatio-temporal space), either analytically by some mathematical formulation (e.g. linear regressors) or by data-driven non-linear learners (e.g. NNs) whereas clustering-based approaches are based on the assumption that the analysis of historical trajectory data provides valid trends and mobility patterns as the basis for TP; current state-of-the-art includes, at least, NN-based [52], statistical data mining [18], vectored area navigation arrivals [41], as well as improvements over Kalman and DBSCAN algorithms [85][27][40][76].
- Specific to the aviation domain, collaborative and other related approaches address the TP task indirectly via the related CD&R and CA problems, in the ATCO context of both pre-flight strategic planning and the online tracking of flights; current state-of-the-art includes, at least, two-phase clustering in the terminal phase of flights [106] and linear functional regression via wavelets for ATCO operations [92].
- Semantic-aware techniques, especially for the challenging task of climb-phase TP, provide a more general and integrative framework for exploiting information from additional dimensions other than the spatio-temporal; current state-of-the-art includes, at least, the Center-TRACON Automation System (CTAS) [83], climb-phase TP for ATM [95][37], as well as probabilistic alternatives via HMMs [6][7].

Table II: Surveyed TP techniques.

| Ref. in this paper | Technique | Input | Output (type) | Dataset(s) used |
|---|---|---|---|---|
| cf. Section 4.1 | HMM by Ayhan and Samet [6] | Trajectories, weather | Trajectory (k-D) {k = 2, 3, 4} | ASDI (FAA), RAP (NOAA) |
| | CTAS software by Gong and McNally [35] | Trajectories (aircraft), weather | Trajectory (4-D) | Ground radar |
| | MOA (Monte Carlo, Viterbi) by Ramos [78] | Trajectories (aircraft), type of aircraft | Trajectory (4-D) | ADS-B |
| cf. Section 4.2 | ANN by Le Fablec et al. [52] | Trajectories (aircraft), aircraft type, RFL | Trajectory | Ground radar |
| | ANN by Cheng et al. [18] | Trajectories (aircraft), historical data | Trajectory (4-D) | Ground radar |
| | Clustering, Regression models by Hong and Lee [41] | Trajectories (aircraft), meteorological data | Trajectory (trajectory pattern) | Ground radar |
| | ANN, GLM, SVR by de Leege et al. [21] | Trajectories (radar tracks / ADS-B), meteorological data | Trajectory | ADS-B, ground radar |
| | Online trajectory clustering by Yang et al. [106] | Trajectories (aircraft), flight attributes data | Trajectory | Ground radar |
| | Multiple regression by Kun and Wei [50] | Trajectories (radar tracks), meteorological data (wind conditions) | Trajectory (4-D) | Ground radar |
| | kNN regression by Tastambekov et al. [92] | Trajectories (aircraft) (radar tracks for a given origin–destination pair) | Trajectory (short/mid-term TP) | Ground radar |
| | IMM, DBSCAN, Kalman by Song et al. [85] | Trajectories (aircraft) - (historical radar tracks) | Trajectory (long-term TP) | Ground radar |
| cf. Section 4.3 | MA, GA by Xiangmin et al. [103] | Trajectories (aircraft) | Trajectory (long-term TP) | ADS-B, IFS (radar) |
| | gPC by Matsuno et al. [57] | Trajectories (aircraft), meteorological data (wind parameters) | Trajectory | Model parameters (synthetic data) |
| | Collaborative – CR/CD&A via traffic & complex network analysis by Manathara and Ghose [55] | Trajectories (aircraft), weather, ATCO (traffic) | Trajectory | Ground radar, weather data |

| Ref. in this paper | Technique | Input | Output (type) | Dataset(s) used |
|---|---|---|---|---|
|  | Collaborative – CR/CD&A via traffic & complex network analysis by Albarker and Rahim [2] | Trajectories (aircraft), weather, ATCO (traffic) | Trajectory | Ground radar, weather data |
|  | IMM by Baek and Bang [8] | Trajectories (aircraft) - ADS-B messages | Trajectory | ADS-B |
| cf. Section 4.4 | Collaborative – CR/CD&A via traffic & complex network analysis by Chen et al. [17] | Trajectories (aircraft), weather, ATCO (traffic) | Trajectory | Ground radar, weather data |
|  | SVM by Di Cicio et al. [23] | Trajectories (aircraft) - (airplane's position, velocity and intended destination) | Trajectory | FlightRadar24.com [30], FlightStats.com [31] |
|  | RMIMM, IBTP by Yepes et al. [109] | ATC regulations, flight plans, pilot intent, weather data, ADS-B messages | Trajectory | ADS-B, weather data, flight plans |
|  | CINTIA trajectory predictor by Swierstra and Green [86] | Trajectories (aircraft) | Trajectory | Model parameters (synthetic data) |
| cf. Section 4.5 | Center-TRACON Automation System (CTAS) by Coppenbarger [20] | Real-time aircraft data (aircraft state, aircraft performance, pilot intent and atmospheric data) | Trajectory | Model parameters & CTAS (synthetic data), ATM databases |
|  | Center-TRACON Automation System (CTAS) by Slattery [83] | Real-time aircraft data (aircraft state, aircraft performance, pilot intent and atmospheric data) | Trajectory | Model parameters & CTAS (synthetic data), ATM databases |
|  | Adaptive-weight by Thipphavong et al. [95] | Trajectories (aircraft) | Trajectory (aircraft climb trajectories) | CTAS (synthetic data) |
|  | BADA - (CMA-ES) by Hadjaz et al. [37] | Trajectories (aircraft) | Trajectory (climbing phase) | Model parameters (synthetic data), ground radar |
|  | HMM (k-NN / DTW) by Ayhan and Same [7] | Trajectories (aircraft), weather | Trajectory (climb phase) | ASDI (FAA), RAP (NOAA) |

# 5. CHALLENGES IN THE ERA OF BIG DATA

In the two tables listing the FLP and TP works surveyed in this paper, cf. Section 3.4 and Section 4.6, respectively, we presented the techniques in a compact form in order to highlight the various approaches and their specifications regarding the input/output of the algorithms. Moreover, these tables bring together information about the different datasets that have been used for evaluation purposes in those works. As it turns out, although the mobility domain is unquestionably a field that contributes to the challenges of the big data era (i.e. the well-known 3 V's, namely volume, velocity, and variety), the proposals so far do not adequately address these challenges, especially when the application in hand tackles more than one V. In the sections that follow, we provide a possible view from the future, by highlighting, on the one hand, the key limitations of the current approaches and, on the other hand, interesting research directions that need to be addressed by the research community.

## 5.1 The challenge of 'volume'

From the discussion provided earlier, it is evident that the largest utilized datasets are in the order of a few million-point records or, respectively, a few thousand trajectories, which is by far lower than what real-world location-sensing applications may collect and store: millions of people are now moving around in the urban area of a city like New York, London or Beijing, with a high percentage of them sharing their position with e.g. Location-Based Service (LBS) and Location-Based Social Networking (LBSN) providers, such as Google and Facebook, respectively; tens of thousands of aircrafts are now flying on the sky globally transmitting their position every second for aviation safety purposes; etc. Two are main bottlenecks that current approaches are not up to this task, given the volume specifications of the Big Data era.

First, a general trend of the state-of-the-art approaches is to base their prediction methodology upon patterns which are extracted by applying a machine learning or data mining algorithm customized for mobility data. However, such algorithms follow centralized approaches and they have not been designed so as to cope with vast amount of data. This is not only a matter of implementation of these algorithms in a big data framework (Hadoop, Spark, Flink, etc.), as it well-known that redesigning a technique so as to be applicable in such frameworks is certainly a less than trivial effort. Recently the mobility data mining community has made progress on discovering movement patterns from vast amount of trajectory data, such as [28], however the challenge is that such approaches should be customized in an appropriate way to facilitate the predictive methodology. In other words, the idea is to follow a similar research direction as the community has done so far. To present a concrete example, as researchers have made possible to utilize a centralized trajectory clustering technique (T-OPTICS [69]) in order to support long-term FLP (MyWay [96], surveyed in Section 3), the same roadmap could be followed, not only to invent big-data-capable pattern extraction techniques, but also for these techniques to have predictive characteristics so they can be incorporated in a predictive analytical approach; a recent approach following this way to address the clustering problem is proposed in [22].

Second, another general trend that has shown its merits in several recent proposals is the adoption of the global/local model approach. This approach uses a mobility pattern extraction technique to learn several local models (e.g. frequent sequential trajectory patterns [33]) and then build a higher-level global model that consolidates the local models in an appropriate way (e.g. T-Pattern-tree in WhereNext [62], surveyed in Section 3). Following such an approach is extremely challenging to be pursued in a big data setting, not only due to that the design of a global model should be big-data-ready itself, but even more importantly because the predictive methodology that operates on the global model should be distributed and parallelizable so as to benefit from it. Again, as a concrete example following the above case and assuming that there is no bottleneck in the discovery of T-Patterns (see first challenge above), the challenge is how we could re-design T-Pattern-tree in a distributed environment and how the actual prediction algorithm would run upon it. In general, inventing synergies among various pattern discovery techniques aiming at new mobility predictive analytics is an interesting research roadmap to be followed in the forthcoming years.

## 5.2 The challenge of 'velocity'

From the presentation of the existing FLP and TP methods, surveyed in Section 3 and Section 4, respectively, it is straightforward to conclude that their primary goal was not to address the online, streaming nature of the input, but to propose reliable prediction methods. To be more specific, a closer look at the various algorithms makes it evident that the computational paradigm followed deviates from the typical window-based, micro-batch streaming techniques that are preferred when real-time predictions are required [3]. One could argue that there is no interesting challenge here as the predictions are made per moving object, thus the problem is trivially parallelizable, and the data of a single moving object is small and easily consumed by an algorithm. It is true that such an approach could scale to large fleets of moving objects. Approaches that are capable of providing real-time location predictions in an online fashion with appropriate implementations on big-data computing platforms have only very recently appeared in the literature [99][100].

However, it has been recognized that the most promising prediction methods, especially in application scenarios where long-term predictions are required, are those that either take into account the context wherein the movement of the objects take place (e.g. the movements of nearby objects [29] or make use of historical data and patterns. To make use of such extra knowledge in an online fashion by concurrently processing high-speed location data implies novel hybrid approaches where the advice coming from the extra information to improve the prediction without it, should be extracted at operational data, before the data or the pattern becomes obsolete. Of course, such an approach further implies the ability to timely update the patterns in an incremental way.

Even more interestingly, in several application domains the key 'velocity' challenge is not necessarily inferred by the extreme speed with which the location information changes but, rather it comes from the fact that there are multiple data sources that monitor the same moving objects. For instance, in the aviation domain, each aircraft may be monitored by primary and secondary surveillance radars, by on-board sensors, etc.; recall e.g. the ADS-B system and its usage in TP [8]. These data sources collect data with different characteristics (e.g. different sampling rates) and speeds, they cover different geographical regions (i.e. the quality of the coverage varies for each data source in different regions), while their clocks are not guaranteed to be synchronized. Thus, the adaptation of the prediction methodologies so as to improve the accuracy of the predictions in a timely way is really challenging in a cross-streaming scenario with such characteristics.

### 5.3 The challenge of 'variety'

Recently, there exist applications where each time-stamped location is enriched by contextual (also called, semantic) annotation about the movement; hence a contextually-enriched trajectory (also, *semantic trajectory*) of a moving object is defined as a sequence of triples ($p_i$, $t_i$, $k_i$) where $k_i$ is an annotation, typically represented as a bag of keywords declaring the associated context. Typical annotations include stop vs. move episodes [70][75], starting versus ending activity points, (distinguishable) change in velocity vector [72], entry in versus exit from a region of interest, etc. In this line, and in compatibility with the problem definitions for FLP and TP presented in Section 2, we could foresee that in the near future FLP/TP definitions will be expanded by their semantic-aware counterparts. As an example, and recalling Fig. 1, the semantic-aware counterpart of TP problem would estimate when and where a plane would reach 'top of climb', 'top of descent', and 'touch down' phases, an extremely important problem, obviously not only in aviation but in other domains as well: when and where a car is expected to reach the ring of a town, when, where and how long a fishery boat is expected to perform a fishing activity, etc.

Orthogonally to the above, the FLP/TP problems may be assisted by related information gathered for this purpose. For instance, it is only very recently that approaches make use of weather data so as to solve the TP problem [6][7]. In the mobility data field and in any application domain, finding actionable insights from the huge mass of data is challenging to integrate multiple data sources, as this may be either new vs. old location data, big or small data (weather data is much bigger than location data), structured or unstructured data (e.g. location data vs. textual data that annotate the former in location-based social networks).

What is even more challenging is when the variety comes not only from the data in its raw form, but from the processing and fusion with various methods. For instance, in [99][100] the variety is introduced by transforming raw trajectory data to trajectory synopses (by a compression method), which are further transformed to higher-level events (by a composite event detection methodology). Fusing this varied transformed data for the sake of improved predictive analytics is a promising research roadmap to follow.

## 6. CONCLUSIONS

In this article, we provided an extensive survey of the research results in the challenging field of predictive analytics in the area of spatial and mobility data. We presented formal definitions for the different variations of the trajectory prediction problem, introduced a taxonomy of solutions proposed so far and provided a review of almost fifty related methods and techniques. From this study, it turns out that current state-of-the-art is still far from addressing big data challenges that have emerged in real-world applications (the 3 V's: volume, velocity, and variety); hence, the domain is open to new contributions.

# REFERENCES

[1]   Agrawal, R. & Srikant, R. (1994) Fast algorithms for mining association rules in large databases. *Proceedings of VLDB*.

[2]   Albarker, B.M., & Rahim, N.A. (2010) Unmanned aircraft collision avoidance system using cooperative agent-based negotiation approach. *International Journal of of Simulation -- Systems, Science & Technology*, 11(5), 1-7.

[3]   Alevizos, E., Skarlatidis, A., Artikis, A., & Paliouras, G. (2017) Probabilistic complex event recognition: a survey. *ACM Computing Surveys*, 50(5), article no. 71.

[4]   Andrienko, G., Andrienko, N., Bak, P., Keim, D., & Wrobel, S. (2013) *Visual analytics of movement*. Springer Science and Business Media.

[5]   Ashbrook, D., & Starner, T. (2003) Using GPS to learn significant locations and predict movement across multiple users. *Personal and Ubiquitous Computing*, 7(5), 275-286.

[6]   Ayhan, S., & Samet, H. (2016) Aircraft trajectory prediction made easy with predictive analytics. *Proceedings of ACM SIGKDD*.

[7]   Ayhan, S., & Samet, H. (2016) Time series clustering of weather observations in predicting climb phase of aircraft trajectories. *Proceedings of IWCTS*.

[8]   Baek, K., & Bang H. (2012) ADS-B based trajectory prediction and conflict detection for air traffic management. *International Journal of of Aeronautical & Space Science*, 13(3), 377-385.

[9]   Barabasi, A. & Albert, R. (1999) Emergence of scaling in random networks. *Science*, 286, 509-512.

[10]  Beckmann, N., Kriegel, H.-P., Scheider, R., & Seeger, B. (1990) The R*-tree: an efficient and robust access method for points and rectangles. *Proceedings of ACM SIGMOD*.

[11]  Bentley, J.L. (1975) Multidimensional binary search trees used for associative searching. *Communications of the ACM*, 18(9), 509-517.

[12]  Bhattacharya, A., & Das, S. (1999) LeZi-update: an information-theoretic approach to track mobile users in PCS networks. *Proceedings of ACM MobiCom*.

[13]  Bianconi, G. & Barabasi, A. (2001) Bose-Einstein condensation in complex networks. *Physical Review Letters*, 86, 5632-5635.

[14]  Bifet, A., Holmes, G., Kirkby, R., & Pfahringer, B. (2010) MOA: Massive online analysis. *Journal of Machine Learning Research*, 11(5), 1601-1604.

[15]  Bresenham, J. (1965) Algorithm for computer control of a digital plotter. *IBM Systems Journal*, 4(1), 25-30.

[16]  Brinkhoff, T. (2002) A framework for generating network based moving objects. *Geoinfomatica*, 6(2), 153-180.

[17]  Chen, X.W., Landry, S.J., & Nof, S.Y. (2011) A framework of enroute air traffic conflict detection and resolution through complex network analysis. *Computers in Industry*, 62(8), 787-794.

[18]  Cheng, T., Cui, D., & Cheng, P. (2003). Data mining for air traffic flow forecasting: a hybrid model of neural network and statistical analysis. *Proceedings of ITSC*.

[19]  Comer, D. (1979) The ubiquitous B-tree. *ACM Computing Surveys*, 11(2), 121-137.

[20]  Coppenbarger, R.A. (1999) En route climb trajectory prediction enhancement using airplane flight-planning information. *Proceedings of AIAA GNC*.

[21]  de Leege, A., van Paassen, M., & Mulder, M. (2013) A machine learning approach to trajectory prediction. *Proceedings of AIAA GNC*.

[22]  Deng, Z., Hu, Y., Zhu, M., Huang, X., & Du, B. (2015) A scalable and fast OPTICS for clustering trajectory big data. Cluster Computing, 18(2), 549-562.

[23]  Di Ciccio, C., van der Aa, H., Cabanillas, C., Mendling, J., & Prescher, J. (2016) Detecting flight trajectory anomalies and predicting diversions infreight transportation. *Decision Support Systems*, 88, 1-17.

[24]  Eagle, N. & Pentland, A. (2006) Reality mining: sensing complex social systems. *Personal and Ubiquitous Computing*, 10(4), 255-268.

[25] Enea, G. & Porretta, M. (2012) A comparison of 4-D trajectory operations envisioned for NextGen and SESAR - Some preliminary findings. *Proceedings of ICAS*.

[26] Erdos, P. & Renyi, A. (1960) On the evolution of random graphs. *Publication of the Mathematical Institute of the Hungarian Academy of Sciences*, 5, 17-61.

[27] Ester, M., Kriegel, H., Sander, J., & Xu, X. (1996) A density-based algorithm for discovering clusters in large spatial databases with noise. *Proceedings of KDD*.

[28] Fan, Q., Zhang, D., Wu, H., & Tan, K.-L. (2016) A general and parallel platform for mining co-movement patterns over large-scale trajectories. *Proceedings of the VLDB Endowment*, 10(4), 313-324.

[29] Fernández, E.C., Cordero, J.M., Vouros, G., et al. (2017) DART: a machine-learning approach to trajectory prediction and demand-capacity balancing. Proceedings of SESAR Innovation Days. Available online at: https://www.sesarju.eu/sesarinnovationdays.

[30] FlightRadar24 Flight Tracker. URL: https://www.flightradar24.com.

[31] FlightStats Flight Tracker. URL: https://www.flightstats.com.

[32] Gaede, V. & Günther, O. (1998) Multidimensional access methods. *ACM Computing Surveys*, 30(2), 170-231.

[33] Giannotti, F., Nanni, M., Pinelli, F., & Pedreschi, D. (2007) Trajectory pattern mining. *Proceedings of ACM SIGKDD*.

[34] Gomes, J., Phua, C., & Krishnaswamy, S. (2013) Where will you go? Mobile data mining for next place prediction. *Proceedings of DaWaK*.

[35] Gong, C., & McNally, D. (2004) A methodology for automated trajectory prediction analysis. *Proceedings of AIAA GNC*.

[36] Guttman, A. (1984) R-trees: a dynamic index structure for spatial searching. *Proceedings of ACM SIGMOD*.

[37] Hadjaz, A., Marceau, G., Saveant, P., & Schoenauer, M. (2012) Online learning for ground trajectory prediction. *CoRR, Vol. abs/1212.3998*.

[38] Hamed, M., Gianazza, D., Serrurier, M., & Durand, N. (2013) Statistical prediction of aircraft trajectory: regression methods vs point-mass model. *Proceedings of ATM*.

[39] Han, J., Pei, J., & Yin, Y. (2000) Mining frequent patterns without candidate generation. *Proceedings of ACM SIGMOD*.

[40] Haykin, S. (1991) *Adaptive Filter Theory* (2/e). Prentice-Hall.

[41] Hong, S., & Lee, K. (2015) Trajectory prediction for vectored area navigation arrivals. *Journal of Aerospace Information Systems*, 12(7) 490-502.

[42] Ishikawa, Y., Tsukamoto, Y., & Kitagawa, H. (2004) Extracting mobility statistics from indexed spatio-temporal datasets. *Proceedings of STDBM*.

[43] Jagadish, H.V. (1990) On indexing line segments. *Proceedings of VLDB*.

[44] Jensen, C., Lin, D., & Ooi, B. (2004) Query and update efficient B+-tree based indexing of moving objects. *Proceedings of VLDB*.

[45] Jeung, H., Liu, Q., Shen, H., & Zhou, X. (2008) A hybrid prediction model for moving objects. *Proceedings of IEEE ICDE*.

[46] Jia, Y., Wang, Y., Jin, X., & Cheng, X. (2016) Location prediction: a temporal-spatial Bayesian model. ACM Transactions on Intelligent Systems and Technology, 7(3), article no. 31.

[47] Keogh, E. & Ratanamahatana, C. (2005) Exact indexing of dynamic time warping. *Knowledge and Information Systems*, 7(3), 358-386.

[48] Knuth, D.E. (1973) *The Art of Computer Programming, vol. 3: Sorting and Searching*. Addison-Wesley.

[49] Kollios, G., Gunopulos, D., & Tsotras, V. (1999) On indexing mobile objects. *Proceedings of ACM PODS*.

[50] Kun, W., & Wei, P. (2008). A 4-D trajectory prediction model based on radar data. *Proceedings of CCC*.

[51] Laurila, J.K., Gatica-Perez, D., Aad, I., Blom, J., Bornet, O., Do, T.M.T., Dousse, O., Eberle, J., & Miettinen, M. (2012) The mobile data challenge: big data for mobile computing research. *Proceedings of Mobile Data Challenge by Nokia Workshop.*

[52] Le Fablec, Y., & Alliot, J.M. (1999) Using neural networks to predict aircraft trajectories. *Proceedings of ICIS.*

[53] Liu, T., Sadler, C.M., Zhang, P., & Martonosi, M. (2004) Implementing software on resource-constrained mobile sensors: experiences with Impala and ZebraNet. *Proceedings of MobiSys.*

[54] Mamoulis, N., Cao, H., Kollios, G., Hadjieleftheriou, M., Tao, Y., & Cheung, D.W. (2004) Mining, indexing, and querying historical spatiotemporal data. *Proceedings of SIGKDD.*

[55] Manathara, J.G., & Ghose, D. (2011) Reactive algorithm for collision avoidance of multiple realistic UAVs. *Aircraft Engineering and Aerospace Technology*, 83(6), 388-396.

[56] Manolopoulos, Y., Nanopoulos, A., Papadopoulos, A.N., & Theodoridis, Y. (2005) *R-trees: Theory and Applications.* Springer.

[57] Matsuno, Y., Tsuchiya, T., Wei, J., Hwang, I., Matayoshi, N. (2015) Stochastic optimal control for aircraft conflict resolution under wind uncertainty. *Aerospace Science and Technology*, 43, 77-88.

[58] MIT Reality Mining Dataset. Reality Commons, MIT Human Dynamics Lab. URL: http://realitycommons.media.mit.edu/realitymining.html.

[59] MOA – Machine Learning from Streams. University of Waikato, New Zealand. URL: https://moa.cms.waikato.ac.nz.

[60] Mobile Data Challenge (MDC) Dataset. IDIAP Dataset Distribution Portal. URL: http://www.idiap.ch/dataset/mdc.

[61] Mondolini, S. (2013) Improved trajectory information for the future flight planning environment. *Proceedings of ATM.*

[62] Monreale, A., Pinelli, F., Trasarti, R., & Giannotti, F. (2009) WhereNext: a location predictor on trajectory pattern mining. *Proceedings of ACM SIGKDD.*

[63] Moon, B., Jagadish, H.V., Faloutsos, C., Saltz, J.H. (2001) Analysis of the clustering properties of the Hilbert space-filling curve. *IEEE Transactions on Knowledge and Data Engineering*, 13(1), 124-141.

[64] Morzy, M. (2006) Prediction of moving object location based on frequent trajectories. *Proceedings of ISCIS.*

[65] Morzy, M. (2007) Mining frequent trajectories of moving objects for location prediction. *Proceedings of MLDM.*

[66] MTA bus time historical data. Metropolitan Transportation Authority (MTA) Portal. URL: http://web.mta.info/developers/MTA-Bus-Time-historical-data.html.

[67] Musialek, B., Munafo, C., Ryan, H., & Paglione, M. (2010) *Literature survey of trajectory predictor technology.* US Department of Transportation, Federal Aviation Administration, 1-69. Available online at: https://acy.tc.faa.gov/data/_uploaded/Publications/SepMgmtResearchSurveyTechNoteFinalDeliveryNov2010.pdf.

[68] Myers, C. & Rabiner, L. (1981) A comparative study of several dynamic time-warping algorithms for connected word recognition. *The Bell System Technical Journal*, 60(7), 1389-1409.

[69] Nanni, M. & Pedreschi, D. (2006) Time-focused clustering of trajectories of moving objects. *Journal of Intelligent Information Systems*, 27, 267–289.

[70] Paparrizos, J. & Gravano, L. (2016) k-shape: Effcient and accurate clustering of time series. SIGMOD Record, 45(1), 69-76.

[71] Parent, C., Spaccapietra, S., Renso, C., Andrienko, G., Andrienko, N., Bogorny, V., Damiani, M.L., Gkoulalas-Divanis, A., Macedo, J., Pelekis, N., Theodoridis, Y., & Yan, Z. (2013) Semantic trajectories modeling and analysis. *ACM Computing Surveys*, 45(4), article no. 42.

[72] Patroumpas, K., Alevizos, E., Artikis, A., Vodas, M., Pelekis, N., & Theodoridis, Y. (2017) Online event recognition from moving vessel trajectories. *Geoinformatica*, 21(2), 389-427.

[73] Pearson, K. (1895) Notes on regression and inheritance in the case of two parents. *Proceedings of the Royal Society of London, 58*, 240-242.

[74] Pelekis, N., & Theodoridis, Y. (2014) *Mobility Data Management and Exploration*. Springer.

[75] Pelekis, N., Sideridis, S., Tampakis, P., & Theodoridis, Y. (2016) Simulating our lifesteps by example. *ACM Transactions on Spatial Algorithms and Systems*, 2(3), 1-39.

[76] Porat, B. (1994) *Digital Processing of Random Signals - Theory and Methods*. Dover Publications.

[77] Rabiner, L.R. (1989) A tutorial on hidden Markov models and selected applications in speech recognition. *Proceedings of the IEEE*, 77(2), 257-286.

[78] Ramos, J.J.F. (2014) Statistical model for aircraft trajectory prediction. Technical Report, *Instituto Superior Tecnico, Lisboa*. Available online at: https://fenix.tecnico.ulisboa.pt/downloadFile/395146457607/Resumo.pdf.

[79] Saltenis, S., Jensen, C., Leutenegger, S., & Lopez, M. (2000) Indexing the positions of continuously moving objects. *Proceedings of ACM SIGMOD*.

[80] Samet, H. (1990) *The Design and Analysis of Spatial Data Structures*. Addison-Wesley.

[81] Shannon, C.E. (1948) A mathematical theory of communication. *The Bell System Technical Journal*, 27(3), 379-423.

[82] Sistla, A.P., Wolfson, O., Chamberlain, S., & Dao, S. (1997) Modeling and querying moving objects. *Proceedings of IEEE ICDE*.

[83] Slattery, R.A. (1995) Terminal area trajectory synthesis for air traffic control automation. *Proceedings of ACC*.

[84] Solomonoff, R. & Rapoport, A. (1951) Connectivity of random nets. *The Bulletin of Mathematical Biophysics*, 13, 107-117.

[85] Song, Y., Cheng, P., & Mu, C. (2012) An improved trajectory prediction algorithm based on trajectory data mining for air traffic management. *Proceedings of IEEE ICIA*.

[86] Swierstra, S., & Green, S.M. (2003) Common trajectory prediction capability for decision support tools. *Proceedings of ATM*.

[87] Tao, Y., & Papadias, D. (2002) Time-parameterized queries in spatio-temporal databases. *Proceedings of ACM SIGMOD*.

[88] Tao, Y., Sun, J., & Papadias, D. (2003) Selectivity estimation for predictive spatio-temporal queries. *Proceedings of IEEE ICDE*.

[89] Tao, Y., & Papadias, D. (2003) Spatial queries in dynamic environments. *ACM Transactions on Database Systems*, 28(2), 101-139.

[90] Tao, Y., Faloutsos, C., Papadias, D., & Liu, B. (2004) Prediction and indexing of moving objects with unknown motion patterns. *Proceedings of ACM SIGMOD*.

[91] Tao, Y., Papadias, D., & Sun, J. (2003) The TPR*-tree: an optimized spatio-temporal access method for predictive queries. *Proceedings of VLDB*.

[92] Tastambekov, K., Puechmorel, S., Delahaye, D., & Rabut, C. (2014) Aircraft trajectory forecasting using local functional regression in Sobolev space. *Transportation research part C: emerging technologies*, 39, 1-22.

[93] Tayeb, J., Ulusoy, Ö., & Wolfson, O. (1998) A quadtree-based dynamic attribute indexing method. *The Computer Journal*, 41(3), 185-200.

[94] Tcha, D. & Yoon, M. (1995) Conduit and cable installation for a centralized network with logical star̄star topology. *IEEE Transactions on Communications*, 43, 958-967.

[95] Thipphavong, D.P., Schultz, C.A., Lee, A.G., Chan, S.H. (2013) Adaptive algorithm to improve trajectory prediction accuracy of climbing aircraft. *Journal of Guidance, Control and Dynamics*, 36(1), 15-24.

[96] Trasarti, R., Guidotti, R., Monreale, A., & Giannotti, F. (2017) MyWay: Location Prediction via mobility profiling. *Information Systems*, 64, 350-367.

[97] Verhein, F., & Chawla, S. (2006) Mining spatio-temporal association rules, sources, sinks, stationary regions and thoroughfares in object mobility databases. *Proceedings of DASFAA*.

[98] Viterbi, A. (1967) Error bounds for convolutional codes and an asymptotically optimum decoding algorithm. *IEEE Transactions on Information Theory*, 13(2), 260-269.

[99] Vouros, G.A., Vlachou, A., Santipantakis, G., et al. (2018) Big data analytics for time critical mobility forecasting: recent progress and research challenges. *Proceedings of EDBT*.

[100] Vouros, G.A., Vlachou, A., Santipantakis, G., et al. (2018) Increasing maritime situation awareness via trajectory detection, enrichment and recognition of events. *Proceedings of W2GIS*.

[101] WEKA – Machine Learning Algorithms for Data Mining Tasks. University of Waikato, New Zealand. URL: https://www.cs.waikato.ac.nz/ml/weka/.

[102] Wu, F., Fu, K., Wang, Y., Xiao, Z., & Fu, X. (2017) A spatial-temporal-semantic neural network algorithm for location prediction on moving objects. *Algorithms*, 10(2), article no. 37.

[103] Xiangmin, G., Xuejun Z., Dong H., Yanbo Z., Ji L., & Jing S. (2014) A strategic flight conflict avoidance approach based on a memetic algorithm. *Chinese Journal of Aeronautics*, 27(1), 93-101.

[104] Xu, B., & Wolfson, O. (2003) Time-series prediction with applications to traffic and moving objects databases. *Proceedings of ACM MobiDE*.

[105] Yang, J., & Hu, M. (2006) TrajPattern: Mining sequential patterns from imprecise trajectories of mobile objects. *Proceedings of EDBT*.

[106] Yang, Y., Zhang, J., & Cai, K.Q. (2015) Terminal-area aircraft intent inference approach based on online trajectory clustering. *The Scientific World Journal*, 2015, article no. 671360.

[107] Yavas, G., Katsaros, D., Ulusoy, Ö., & Manolopoulos, Y. (2005) A data mining approach for location prediction in mobile environments. *Data and Knowledge Engineering*, 54(2), 121-146.

[108] Ye, Y., Zheng, Y., Chen, Y., Feng, J., & Xie, X. (2009) Mining individual life pattern based on location history. *Proceedings of MDM*.

[109] Yepes, J.L., Hwang, I., & Rotea, M. (2007) New algorithms for aircraft intent inference and trajectory prediction. *Journal of Guidance, Control, and Dynamics*, 30(2), 370-382.

[110] Ying, J.-C., Lee, W.-C., Weng, T.-C., & Tseng, V. (2011) Semantic trajectory mining for location prediction. *Proceedings of ACM SIGSPATIAL*.

[111] Zheng, Y. (2015) Trajectory data mining: an overview. *Transactions on Intelligent Systems and Technology*, 6(3), 1-41.

[112] Zheng, Y., Zhang, L., Xie, X., & Ma, W.-Y. (2009) Mining correlation between locations using human location history. *Proceedings of ACM SIGSPATIAL*.

[113] Zheng, Y., Fu, H., Xie, X., Ma, W.-Y., & Li, Q. (2011) Geolife GPS trajectory dataset – user guide. *Microsoft Research*. Available online at: https://www.microsoft.com/en-us/research/publication/geolife-gps-trajectory-dataset-user-guide/.

[114] Ziv, J. & Lempel, A. (1978) Compression of individual sequences via variable-rate coding. *IEEE Transactions on Information Theory, 24(5), 530–536*.

[115] Zorbas, N., Zissis, D., Tserpes, K., & Anagnostopoulos, D. (2015) Predicting object trajectories from high-speed streaming data. *Proceedings of IEEE TrustCom/BigDataSE/ISPA*.

[116] Hendawi, A., Ali, M. & Mokbel, M. (2015) A Framework for Spatial Predictive Query Processing and Visualization. *16th IEEE International Conference on Mobile Data Management*.

[117] Zhang, R., Qi, J., Lin, D., Wang, W. & Chi-WingWong, R. (2012) A highly optimized algorithm for continuous intersection join queries over moving objects, *VLDB Journal, 21, 561–586*.

[118] Zhang, M., Chen, S., Jensen, C., Ooi, B. & Zhang, Z. (2009) Effectively Indexing Uncertain Moving Objects for Predictive Queries, *Proceedings of VLDB '09, August 24-28, 2009, Lyon, France.*

[119] Hendawi, A., Bao, J., Mokbel & M. (2013) iRoad: A Framework For Scalable Predictive Query Processing on Road Networks, *Proceedings of VLDB'13, August 26-30, 2013, Trento, Italy.*

[120] Hendawi, A. & Mokbel, M. (2012) Panda: A Predictive Spatio-Temporal Query Processor, *Proceedings of ACM SIGSPATIAL GIS'12, November 6-9, 2012, CA, USA.*

[121] Jeung, H., Yiu, M., Zhou, X. & Jensen, C. (2010) Path prediction and predictive range querying in road network databases, *VLDB Journal, 19, 585–602.*

[122] Hendawi, A. & Mokbel, M. (2012) Predictive Spatio-Temporal Queries: A Comprehensive Survey and Future Directions, *Proceedings of ACM SIGSPATIAL MobiGIS'12, November 6, 2012, CA, USA.*

[123] Hendawi, A., Bao, J., Mokbel, M. & Ali, M. (2015) Predictive Tree: An Efficient Index for Predictive Queries on Road Networks, *Proceedings of ICDE 2015.*

[124] Ding, X., Chenz, L., Gao, Y., Jensenz, C. & Bao, H. (2018) UlTraMan: A Unified Platform for Big Trajectory Data Management and Analytics, *Proceedings of VLDB'18, August 2018, Rio de Janeiro, Brazil.*

[125] Hendawi, A., Ali, M. & Mokbel, M. (2017) Panda *: A generic and scalable framework for predictive spatio-temporal queries, *GeoInformatica, April 2017, 21(2), 175–208.*

[126] Millefiori, L., Braca, P., Bryan, K. & Willett, P. (2016) Modeling Vessel Kinematics Using a Stochastic Mean-Reverting Process for Long-Term Prediction, *IEEE Trans. on Aerospace and Electronic Systems, 52(5), 2313-2330.*

[127] Vivone, G., Millefiori, L., Braca & Willett, P. (2017) Performance Assessment of Vessel Dynamic Models for Long-Term Prediction Using Heterogeneous Data, *IEEE Trans. on Geoscience and Remote Sensing, 55(11), 6533-6546.*